\RequirePackage[loading]{tracefnt}
\documentclass{article}
\usepackage{times}
\usepackage{graphicx}
\usepackage{amsmath,amssymb,enumerate,bbm,color,alltt}
\usepackage{hyperref}
\usepackage{url}
\usepackage{graphicx}
\usepackage{float}
\usepackage{stfloats}
\usepackage{subfig}
\usepackage{epstopdf}
\usepackage{longtable}
\usepackage{tabularx}
\usepackage{multirow}
\usepackage{booktabs}
\usepackage{subfig}
\usepackage[titletoc]{appendix}

\newcommand{\nobracket}{}

\newcommand{\tmmathbf}[1]{\ensuremath{\boldsymbol{#1}}}

\newcommand{\tmop}[1]{\ensuremath{\operatorname{#1}}}

\newenvironment{enumeratenumeric}{\begin{enumerate}[1.] }{\end{enumerate}}

\usepackage[accepted]{icml2016}

\icmltitlerunning{Neural Variational Inference for Text Processing}

\begin{document} 
\setlength{\abovedisplayskip}{6pt}
\setlength{\belowdisplayskip}{6pt}
\twocolumn[
\icmltitle{Neural Variational Inference for Text Processing}
\icmlauthor{Yishu Miao$^1$}{yishu.miao@cs.ox.ac.uk}
\icmlauthor{Lei Yu$^1$}{lei.yu@cs.ox.ac.uk}
\icmlauthor{Phil Blunsom$^{12}$}{phil.blunsom@cs.ox.ac.uk}
\icmladdress{$^1$University of Oxford, $^2$Google Deepmind}

\icmlkeywords{boring formatting information, machine learning, ICML}

\vskip 0.3in
]

\begin{abstract}
Recent advances in neural variational inference have spawned a renaissance in deep latent variable models.
In this paper we introduce a generic variational inference framework for generative and conditional models of text.
While traditional variational methods derive an analytic approximation for the intractable distributions over latent variables, here we construct an inference network conditioned on the discrete text input to provide the variational distribution.
We validate this framework on two very different text modelling applications, generative document modelling and supervised question answering.
Our neural variational document model combines a continuous stochastic document representation with a bag-of-words generative model and achieves the lowest reported perplexities on two standard test corpora.
The neural answer selection model employs a stochastic representation layer within an attention mechanism to extract the semantics between a question and answer pair. 
On two question answering benchmarks this model exceeds all previous published benchmarks.

\end{abstract}

\section{Introduction}

\begin{figure*}[htb]
\vspace{-0.5em}
\begin{minipage}[t]{0.5\linewidth}
  \centering
	\includegraphics[width=1.7in]{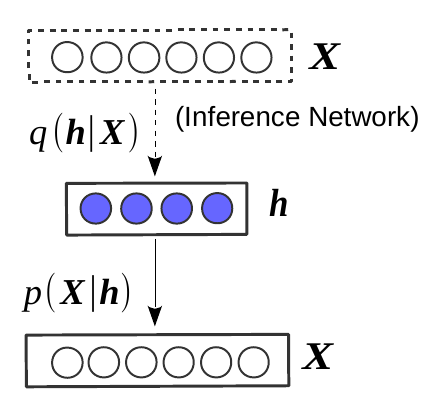}
	\vspace{-1em}
\caption{NVDM for document modelling.}
\label{fig:vtm}
\end{minipage}%
\begin{minipage}[t]{0.5\linewidth}
  \centering
	\includegraphics[width=3.2in]{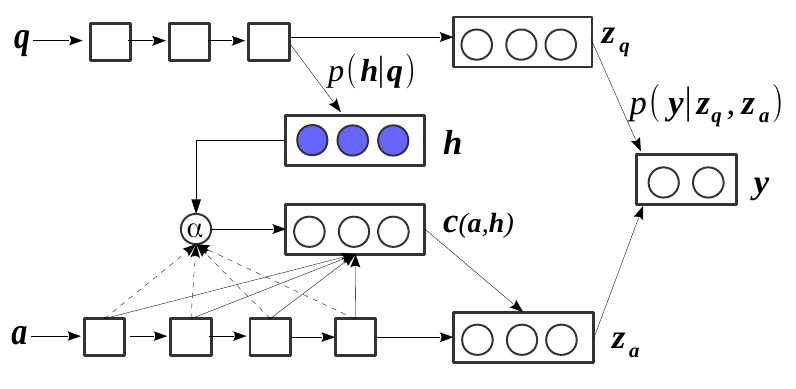}
	\vspace{-1em}
\caption{NASM for question answer selection.}
\label{fig:vsm}
\end{minipage}
\vspace{-1em}
\end{figure*}

Probabilistic generative models underpin many successful applications within the field of natural language processing (NLP). 
Their popularity stems from their ability to use unlabelled data effectively, to incorporate abundant linguistic features, and to learn interpretable dependencies among data.
However these successes are tempered by the fact that as the structure of such generative models becomes deeper and more complex, true Bayesian inference becomes intractable due to the high dimensional integrals required. 
Markov chain Monte Carlo (MCMC) \citep{neal1993probabilistic,andrieu2003introduction} and variational inference \citep{jordan1999introduction,attias2000variational,beal2003variational} are the standard approaches for approximating these integrals. 
However the computational cost of the former results in impractical training for the large and deep neural networks which are now fashionable, and the latter is conventionally confined due to the underestimation of posterior variance.
The lack of effective and efficient inference methods hinders our ability to create highly expressive models of text, especially in the situation where the model is non-conjugate. 

This paper introduces a neural variational framework for generative models of text, inspired by the variational auto-encoder \citep{rezende2014stochastic,kingma2013auto}. 
The principle idea is to build an inference network, implemented by a deep neural network conditioned on text, to approximate the intractable distributions over the latent variables. 
Instead of providing an analytic approximation, as in traditional variational Bayes, neural variational inference learns to model the posterior probability, thus endowing the model with strong generalisation abilities. 
Due to the flexibility of deep neural networks, the inference network is capable of learning complicated non-linear distributions and processing structured inputs such as word sequences. 
Inference networks can be designed as, but not restricted to, multilayer perceptrons (MLP), convolutional neural networks (CNN), and recurrent neural networks (RNN), approaches which are rarely used in conventional generative models.
By using the reparameterisation method \citep{rezende2014stochastic,kingma2013auto}, the inference network is trained through back-propagating unbiased and low variance gradients w.r.t.\ the latent variables.
Within this framework, we propose a Neural Variational Document Model (NVDM) for document modelling and a Neural Answer Selection Model (NASM) for question answering, a task that selects the sentences that correctly answer a factoid question from a set of candidate sentences.

The NVDM (Figure \ref{fig:vtm}) is an unsupervised generative model of text which aims to extract a continuous semantic latent variable for each document. 
This model can be interpreted as a variational auto-encoder: an MLP encoder (inference network) compresses the bag-of-words document representation into a continuous latent distribution, and a softmax decoder (generative model) reconstructs the document by generating the words independently. 
A primary feature of NVDM is that each word is generated directly from a dense continuous document representation instead of the more common binary semantic vector \citep{hinton2009replicated,larochelle2012neural,Srivastava2013,mnih2014neural}.
Our experiments demonstrate that our neural document model achieves the state-of-the-art perplexities on the \textit{20NewsGroups} and \textit{RCV1-v2}.

The NASM (Figure \ref{fig:vsm}) is a supervised conditional model which imbues LSTMs \citep{hochreiter1997long} with a latent stochastic attention mechanism to model the semantics of question-answer pairs and predict their relatedness. 
The attention model is designed to focus on the phrases of an answer that are strongly connected to the question semantics and is modelled by a latent distribution.
This mechanism allows the model to deal with the ambiguity inherent in the task and learns pair-specific representations that are more effective at predicting answer matches, rather than independent embeddings of question and answer sentences. 
Bayesian inference provides a natural safeguard against overfitting, especially as the training sets available for this task are small.
The experiments show that the LSTM with a latent stochastic attention mechanism learns an effective attention model and outperforms both previously published results, and our own strong non-stochastic attention baselines.

In summary, we demonstrate the effectiveness of neural variational inference for text processing on two diverse tasks. These models are simple, expressive and can be trained efficiently with the highly scalable stochastic gradient back-propagation.
Our neural variational framework is suitable for both unsupervised and supervised learning tasks, and can be generalised to incorporate any type of neural networks.

\section{Neural Variational Inference Framework}
Latent variable modelling is popular in many NLP problems, but it is non-trivial to carry out effective and efficient inference for models with complex and deep structure. In this section we introduce a generic neural variational inference framework that we apply to both the unsupervised NVDM and supervised NASM in the follow sections. 

We define a generative model with a latent variable $\tmmathbf{h}$, which can be considered as the stochastic units in deep neural networks. We designate the observed parent and child nodes of $\tmmathbf{h}$ as $\tmmathbf{x}$ and $\tmmathbf{y}$ respectively. Hence, the joint distribution of the generative model is $p_{\theta}( \tmmathbf{x}, \tmmathbf{y})= \sum_{\tmmathbf{h}} p_{\theta}(\tmmathbf{y}|\tmmathbf{h}) p_{\theta}(\tmmathbf{h}|\tmmathbf{x}) p(\tmmathbf{x}) $, and the variational lower bound $\mathcal{L}$ is derived as:
\begin{align}
  \mathcal{L} & = \mathbb{E}_{q_{ } ( \tmmathbf{h})}  [\log p_{\theta}
  ( \tmmathbf{y} | \tmmathbf{h}) p_{\theta} ( \tmmathbf{h} |
  \tmmathbf{\tmmathbf{x}}) p ( \tmmathbf{x}) - \log q_{ } (
  \tmmathbf{h})] \label{eq:lb_framework}\\
			& \leqslant \log \int \frac{q_{ } ( \tmmathbf{h})}{q_{ } ( \tmmathbf{h})}
  p_{\theta} ( \tmmathbf{y} | \tmmathbf{h}) p_{\theta} ( \tmmathbf{h} |
  \tmmathbf{x}) p ( \tmmathbf{x}) d \tmmathbf{h} = \log p_{\theta} ( \tmmathbf{x}, \tmmathbf{y}) \nonumber   
\end{align}
where $\theta$ parameterises the generative distributions $ p_{\theta} ( \tmmathbf{y} | \tmmathbf{h})$  and $p_{\theta} ( \tmmathbf{h} | \tmmathbf{\tmmathbf{x}})$. 
In order to have a tight lower bound, the variational distribution $q_{ } ( \tmmathbf{h})$ should approach the true posterior $p ( \tmmathbf{h} | \tmmathbf{\tmmathbf{x}}, \tmmathbf{y})$.
Here, we employ a parameterised diagonal Gaussian  $\mathcal{N} ( \tmmathbf{h} | \tmmathbf{\mu} (\tmmathbf{\tmmathbf{x}}, \tmmathbf{y}), \mathrm{diag} ( \tmmathbf{\sigma}^2 (\tmmathbf{\tmmathbf{x}}, \tmmathbf{y})))$ as $q_{\phi} ( \tmmathbf{h} | \tmmathbf{\tmmathbf{x}}, \tmmathbf{y})$. The three steps to construct the inference network are:
\begin{enumeratenumeric}
 \setlength\itemsep{-0.01cm}
  \item Construct vector representations of the observed variables:
  $\tmmathbf{u} = f_x ( \tmmathbf{x})$, $\tmmathbf{v} = f_y ( \tmmathbf{y})$.
  \item Assemble a joint representation:
  $\tmmathbf{\pi} = g ( \tmmathbf{u}, \tmmathbf{v})$.
  \item Parameterise the variational distribution over the latent variable:
  $\tmmathbf{\mu} = l_1 ( \tmmathbf{\pi}), \log \tmmathbf{\sigma} = l_2 (
  \tmmathbf{\pi})$.
\end{enumeratenumeric}
$f_x (\cdot)$ and $f_y (\cdot)$ can be any type of deep neural networks that are suitable for the observed data; $g (\cdot)$ is an MLP that concatenates the vector representations of the conditioning variables; $l(\cdot)$ is a linear transformation which outputs the parameters of the Gaussian distribution. 
By sampling from the variational distribution, $\tmmathbf{h} \sim q_{\phi} ( \tmmathbf{h} | \tmmathbf{\tmmathbf{x}}, \tmmathbf{y})$, we are able to carry out stochastic back-propagation to optimise the lower bound (Eq. \ref{eq:lb_framework}). 

During training, the model parameters $\theta$ together with the inference network parameters $\phi$ are updated by stochastic back-propagation based on the samples $\tmmathbf{h}$ drawn from $q_{\phi} ( \tmmathbf{h} | \tmmathbf{\tmmathbf{x}}, \tmmathbf{y})$. 
For the gradients w.r.t.\ $\theta$, we have the form:
\begin{eqnarray}
  & \nabla_{\theta}  \mathcal{L} \simeq \frac{1}{L}  \sum_{l = 1}^L
  \nabla_{\theta} \log p_{\theta} ( \tmmathbf{y} | \tmmathbf{h}^{(l)})
  p_{\theta} ( \tmmathbf{h}^{(l)} | \tmmathbf{\tmmathbf{x}}) & 
\end{eqnarray}
For the gradients w.r.t.\ $\phi$ we reparameterise $\tmmathbf{h} =\tmmathbf{\mu} + \tmmathbf{\sigma} \cdot \tmmathbf{\epsilon}$ and sample $\tmmathbf{\epsilon}^{(l)} \sim \mathcal{N} (0, \tmmathbf{I})$ to reduce the variance in stochastic estimation  {\citep{rezende2014stochastic,kingma2013auto}}. 
The update of $\phi$ can be carried out by back-propagating the gradients w.r.t.\ $\tmmathbf{\mu}$ and $\tmmathbf{\sigma}$:
\begin{eqnarray}
　\nonumber　&　 s(\tmmathbf{h})  = \log p_{\theta} ( \tmmathbf{y} | \tmmathbf{h})
  p_{\theta} ( \tmmathbf{h}| \tmmathbf{\tmmathbf{x}}) 
   - \log q_{\phi} (  \tmmathbf{h} | \tmmathbf{\tmmathbf{x}}, \tmmathbf{y}) &\\
   &\nabla_{\mu}  \mathcal{L}  \simeq \frac{1}{L}  \sum_{l = 1}^L
  \nabla_{\tmmathbf{h}^{(l)}}  [ s(\tmmathbf{h}^{(l)}) ]& \\
  & \nabla_{\sigma}  \mathcal{L} \simeq \frac{1}{2 L}  \sum_{l = 1}^L
  \tmmathbf{\epsilon}^{(l)} \nabla_{\tmmathbf{h}^{(l)}}  [s(\tmmathbf{h}^{(l)})] & 
\end{eqnarray}
It is worth mentioning that unsupervised learning is a special case of the neural variational framework where $\tmmathbf{h}$ has no parent node $\tmmathbf{x}$. In that case $\tmmathbf{h}$ is directly drawn from the prior $p(\tmmathbf{h})$ instead of the conditional distribution $p_{\theta}(\tmmathbf{h}|\tmmathbf{x})$, and $s(\tmmathbf{h})  = \log p_{\theta} ( \tmmathbf{y} | \tmmathbf{h}) p_{\theta} (\tmmathbf{h})-\log q_{\phi} ( \tmmathbf{h}|\tmmathbf{y})$.

Here we only discuss the scenario where the latent variables are continuous and the parameterised diagonal Gaussian is employed as the variational distribution. 
However the framework is also suitable for discrete units, and the only modification needed is to replace the Gaussian with a multinomial parameterised by the outputs of a softmax function. 
Though the reparameterisation trick for continuous variables is not applicable for this case, a policy gradient approach {\citep{mnih2014neural}} can help to alleviate the high variance problem during stochastic estimation. \cite{kingma2014semi} proposed a variational inference framework for semi-supervised learning, but the prior distribution over the hidden variable $p(\tmmathbf{h})$ remains as the standard Gaussian prior, while we  apply a conditional parameterised Gaussian distribution, which is jointly learned with the variational distribution.

\section{Neural Variational Document Model}
\label{sec:vtm}\label{sec:vtm}The Neural Variational Document Model (Figure \ref{fig:vtm}) is a simple instance of unsupervised learning  where a continuous hidden variable $\tmmathbf{h} \in \mathbb{R}^K$, which generates all the words in a document independently, is introduced to represent its semantic content. 
Let $\tmmathbf{X} \in \mathbb{R}^{|V|}$ be the bag-of-words representation of a document and $\tmmathbf{x}_i\in \mathbb{R}^{|V|}$ be the one-hot representation of the word at position $i$. 

As an unsupervised generative model, we could interpret NVDM as a variational autoencoder: an MLP encoder $q(\tmmathbf{h}|\tmmathbf{X})$ compresses document representations into continuous hidden vectors ($\tmmathbf{X} \to \tmmathbf{h}$); a softmax decoder $p(\tmmathbf{X}|\tmmathbf{h})=\prod_{i=1}^N p(\tmmathbf{x}_i|\tmmathbf{h})$ reconstructs the documents by independently generating the words ($\tmmathbf{h} \to \{\tmmathbf{x}_i\}$). 
To maximise the log-likelihood $\log \sum\nolimits_{\tmmathbf{h}} p(\tmmathbf{X}|\tmmathbf{h})p(\tmmathbf{h})$ of documents, we derive the lower bound:
\begin{align}
   \mathcal{L}\!\!  = & \mathbb{E}_{q_{\phi}\! (
  \tmmathbf{h} | \tmmathbf{X})} \!  \left[\! \sum_{i = 1}^N \! \log p_{\theta}\!
  ( \tmmathbf{x}_{i} | \tmmathbf{\tmmathbf{h}}) \!\right] \!\!  -  \!\! D_{\tmop{KL}} [q_{\phi} \! ( \tmmathbf{h} | \tmmathbf{X}\!)\|p (
  \tmmathbf{h})  ]  \label{eq:lb}
\end{align}
where $N$ is the number of words in the document and $p(\tmmathbf{h})$ is a Gaussian prior for $\tmmathbf{h}$. 
Here, we consider $N$ is observed for all the documents.
The conditional probability over words $p_\theta(\tmmathbf{x}_i|\tmmathbf{h})$ (decoder) is modelled by multinomial logistic regression and shared across documents:
\begin{equation}
  p_{\theta} ( \tmmathbf{x}_i | \tmmathbf{h}  ) = \frac{\exp \{- E (
  \tmmathbf{x}_i ; \tmmathbf{h}, \theta))\}}{\sum_{j = 1}^{|V|} \exp \{- E (
  \tmmathbf{x}_j ; \tmmathbf{h}, \theta) \}}
  \vspace{-0.5em}
\end{equation}
\begin{equation}
  E ( \tmmathbf{x}_i ; \tmmathbf{h}, \theta) = - \tmmathbf{h}^T \tmmathbf{R}
  \tmmathbf{x}_i - \tmmathbf{b}_{x_i}
\end{equation}
where $\tmmathbf{R} \in \mathbb{R}^{K \times |V|}$ learns the semantic word embeddings and $\tmmathbf{b}_{x_i}$ represents the bias term.

As there is no supervision information for the latent semantics, $\tmmathbf{h}$, the posterior approximation $q_{\phi} ( \tmmathbf{h} | \tmmathbf{X})$ is only conditioned on the current document $\tmmathbf{X}$. 
The inference network $q_{\phi} ( \tmmathbf{h} | \tmmathbf{X})=\mathcal{N} ( \tmmathbf{h} | \tmmathbf{\mu} (\tmmathbf{X}), diag(\tmmathbf{\sigma}^2 (\tmmathbf{X})))$ is modelled as:
\begin{eqnarray}
\label{eq:mlp}
  &\tmmathbf{\pi}  =  g(f^{\textsc{mlp}}_X (\tmmathbf{X})) & \\ 
  &\tmmathbf{\mu}  =  l_1 ( \tmmathbf{\pi}) , 
  \log \tmmathbf{\sigma}  =  l_2 ( \tmmathbf{\pi}) &
\end{eqnarray}
For each document $\tmmathbf{X}$, the neural network generates its own parameters $\tmmathbf{\mu}$ and $\tmmathbf{\sigma}$ that parameterise the latent distribution over document semantics $\tmmathbf{h}$. Based on the samples $\tmmathbf{h} \sim q_{\phi} ( \tmmathbf{h}|\tmmathbf{X})$, the lower bound (Eq. \ref{eq:lb}) can be optimised by back-propagating the stochastic gradients w.r.t. $\theta$ and $\phi$.

Since $p (\tmmathbf{h}  )$ is a standard Gaussian prior, the Gaussian KL-Divergence $D_{\tmop{KL}} [q_{\phi} ( \tmmathbf{h} | \tmmathbf{X})\|p (\tmmathbf{h} )]$ can be computed analytically to further lower the variance of the gradients.
Moreover, it also acts as a regulariser for updating the parameters of the inference network $q_{\phi} ( \tmmathbf{h}| \tmmathbf{X})$.

\section{Neural Answer Selection Model}
Answer sentence selection is a question answering paradigm where a model must identify the correct sentences answering a factual question from a set of candidate sentences. 
Assume a question $\tmmathbf{q}$ is associated with a set of answer sentences $\{\tmmathbf{a}_1,\tmmathbf{a}_2,...,\tmmathbf{a}_n\}$, together with their judgements $\{\tmmathbf{y}_1,\tmmathbf{y}_2,...,\tmmathbf{y}_n\}$, where $\tmmathbf{y}_m=1$ if the answer $\tmmathbf{a}_m$ is correct and $\tmmathbf{y}_m=0$ otherwise. 
This is a classification task where we treat each training data point as a triple $(\tmmathbf{q},\tmmathbf{a},\tmmathbf{y})$ while predicting $\tmmathbf{y}$ for the unlabelled question-answer pair $(\tmmathbf{q},\tmmathbf{a})$.  

The Neural Answer Selection Model (Figure \ref{fig:vsm}) is a supervised model that learns the question and answer representations and predicts their relatedness.
It employs two different LSTMs to embed raw question inputs $\tmmathbf{q}$ and answer inputs $\tmmathbf{a}$. 
Let $\tmmathbf{s}_q (j)$ and $\tmmathbf{s}_a (i)$ be the state outputs of the two LSTMs, and $i$, $j$ be the positions of the states. 
Conventionally, the last state outputs $\tmmathbf{s}_q (| \tmmathbf{q} |)$ and $\tmmathbf{s}_a (|\tmmathbf{a} |)$, as the independent question and answer representations, can be used for relatedness prediction. 
In NASM, however, we aim to learn pair-specific representations through a latent attention mechanism, which is more effective for pair relatedness prediction.

NASM applies an attention model to focus on the words in the answer sentence that are prominent for predicting the answer matched to the current question. 
Instead of using a deterministic question vector, such as $\tmmathbf{s}_q (| \tmmathbf{q} |)$, NASM employs a latent distribution $p_\theta (\tmmathbf{h}|\tmmathbf{q})$ to model the question semantics, which is a parameterised diagonal Gaussian $\mathcal{N}(\tmmathbf{h}|\tmmathbf{\mu}(\tmmathbf{q}),  \mathrm{diag} ( \tmmathbf{\sigma}^2 ( \tmmathbf{q})))$. 
Therefore, the attention model extracts a context vector $\tmmathbf{c}(\tmmathbf{a},\tmmathbf{h})$ by iteratively attending to the answer tokens based on the stochastic vector $\tmmathbf{h} \sim p_\theta (\tmmathbf{h}|\tmmathbf{q}) $. 
In doing so the model is able to adapt to the ambiguity inherent in questions and obtain salient information through attention. 
Compared to its deterministic counterpart (applying $\tmmathbf{s}_q (| \tmmathbf{q} |)$ as the question semantics), the stochastic units incorporated into NASM allow multi-modal attention distributions. 
Further, by marginalising over the latent variables, NASM is more robust against overfitting, which is important for small question answering training sets.

In this model, the conditional distribution $p_\theta (\tmmathbf{h}|\tmmathbf{q})$ is:
\begin{eqnarray}
 & \tmmathbf{\pi}_\theta  =  g_\theta(f^{\textsc{lstm}}_q ( \tmmathbf{q})) = g_\theta(\tmmathbf{s}_q (| \tmmathbf{q} |)) & \label{eq:mlp2} \\ 
 & \tmmathbf{\mu}_\theta  =  l_1 ( \tmmathbf{\pi}_\theta),
  \log \tmmathbf{\sigma}_\theta  =  l_2 ( \tmmathbf{\pi}_\theta) &
\end{eqnarray}
For each question $\tmmathbf{q}$, the neural network generates the corresponding parameters $\tmmathbf{\mu}$ and $\tmmathbf{\sigma}$ that parameterise the latent distribution over question semantics $\tmmathbf{h}$. Following \citet{bahdanau2014neural}, the attention model is defined as:
\begin{eqnarray}
  \alpha (i) & \!\!\! \propto \!\!\! & \exp ( \tmmathbf{W}_{\alpha}^T  \tanh ( \tmmathbf{W}_h  \tmmathbf{h} + \tmmathbf{W}_s \tmmathbf{s}_a (i))) \qquad \\
  \tmmathbf{c}(\tmmathbf{a},\tmmathbf{h}) &\!\!\! = \!\!\! &  \sum\nolimits_i \tmmathbf{s}_a (i)
  \alpha (i)  \label{eq:c} \\
  \tmmathbf{z}_a(\tmmathbf{a},\tmmathbf{h}) & \!\!\! = \!\!\!& \tanh \left( \tmmathbf{W}_a \tmmathbf{c}(\tmmathbf{a},\tmmathbf{h})  + \tmmathbf{W}_n  \tmmathbf{s}_a (| \tmmathbf{a} |) \right) \label{eq:za}
\end{eqnarray}
where $\alpha (i)$ is the normalised attention score at answer token $i$, and the context vector $\tmmathbf{c}(\tmmathbf{a},\tmmathbf{h})$ is the weighted sum of all the state outputs $\tmmathbf{s}_a (i)$. 
We adopt $\tmmathbf{z}_q ( \tmmathbf{q}),\tmmathbf{z}_a ( \tmmathbf{a},\tmmathbf{h})$ as the question and answer representations for predicting their relatedness $\tmmathbf{y}$. 
$\tmmathbf{z}_q ( \tmmathbf{q})$ is a deterministic vector that is equal to $\tmmathbf{s}_q (| \tmmathbf{q} |)$, while $\tmmathbf{z}_a ( \tmmathbf{a}, \tmmathbf{h})$ is a combination of the sequence output $\tmmathbf{s}_a (| \tmmathbf{\tmmathbf{a}}|)$ and the context vector $\tmmathbf{c}(\tmmathbf{a},\tmmathbf{h})$ (Eq. \ref{eq:za}). 
For the prediction of pair relatedness $\tmmathbf{y}$, we model the conditional probability distribution $p_{\theta}  (\tmmathbf{y}| \nobracket
\tmmathbf{z}_q, \tmmathbf{z}_a)$ by sigmoid function:
\begin{eqnarray}\label{qa_match}
  & p_{\theta} (\tmmathbf{y}= 1| \tmmathbf{z}_q, \tmmathbf{z}_a \nobracket) =
  \sigma \left( \tmmathbf{z}_q^T \mathbf{M} \tmmathbf{z}_a + b \right) & 
\end{eqnarray}
To maximise the log-likelihood $\log p (\tmmathbf{y}| \tmmathbf{q}, \tmmathbf{a})$ we use the variational lower bound:
\begin{align}
  \nonumber \mathcal{L}\!\! = & \mathbb{E}_{q_{\phi}\! ( \tmmathbf{h})}\! [\log p_{\theta}
  (\tmmathbf{y}| \tmmathbf{z}_q\! ( \tmmathbf{q}), \tmmathbf{z}_a \!(
  \tmmathbf{a}, \tmmathbf{h}) \nobracket)] \!\! -\!\! D_{\tmop{KL}}\!  (q_{\phi}\! ( \tmmathbf{h} \nobracket) ||p_{\theta}\! ( \tmmathbf{h} | \tmmathbf{q})) \\
  \nonumber \leqslant & \log
  \int p_{\theta} (\tmmathbf{y}| \tmmathbf{z}_q (
  \tmmathbf{q}), \tmmathbf{z}_a ( \tmmathbf{a}, \tmmathbf{h})) p_{\theta} (
  \tmmathbf{h} | \tmmathbf{q}) d\tmmathbf{h}\\
  = & \log p (\tmmathbf{y}| \tmmathbf{q}, \tmmathbf{a}) \label{eq:lb_nasm}
\end{align}
Following the neural variational inference framework, we construct a deep neural network as the inference network $q_{\phi} ( \tmmathbf{h} | \tmmathbf{q}, \tmmathbf{a}, \tmmathbf{y})=\mathcal{N}(\tmmathbf{h}|\tmmathbf{\mu}_\phi(\tmmathbf{q},\tmmathbf{a},\tmmathbf{y}),  \mathrm{diag} ( \tmmathbf{\sigma}^2_\phi ( \tmmathbf{q},\tmmathbf{a},\tmmathbf{y})))$:
\begin{eqnarray}
   \nonumber &\tmmathbf{\pi}_\phi  =  g_\phi ( f^\textsc{lstm}_q(\tmmathbf{q}), f^\textsc{lstm}_a(\tmmathbf{a}),f_y(\tmmathbf{y})) &\\
   &= g_\phi(\tmmathbf{s}_q (| \tmmathbf{q} |),\tmmathbf{s}_a (| \tmmathbf{a} |),\tmmathbf{s}_y) \quad \  & \label{eq:jointMLP} \\
  &\tmmathbf{\mu}_\phi  =  l_3 ( \tmmathbf{\pi}_\phi) ,
  \log \tmmathbf{\sigma}_\phi  =  l_4 ( \tmmathbf{\pi}_\phi)&
\end{eqnarray}
where $\tmmathbf{q}$ and $\tmmathbf{a}$ are also modelled by LSTMs\footnote{In this case, the LSTMs for $\tmmathbf{q}$ and $\tmmathbf{a}$ are shared by the inference network and the generative model, but there is no restriction on using different LSTMs in the inference network.}, and the relatedness label $\tmmathbf{y}$ is modelled by a simple linear transformation into the vector $\tmmathbf{s}_y$. 
According to the joint representation $\tmmathbf{\pi}_\phi$, we then generate the parameters $\tmmathbf{\mu}_\phi$ and $\tmmathbf{\sigma}_\phi$, which parameterise the variational distribution over the question semantics $\tmmathbf{h}$. 
To emphasise, though both $p_{\theta}(\tmmathbf{h} |\tmmathbf{q})$ and $q_{\phi}(\tmmathbf{h} |\tmmathbf{q},\tmmathbf{a},\tmmathbf{y})$ are modelled as parameterised Gaussian distributions, $q_{\phi}(\tmmathbf{h} |\tmmathbf{q},\tmmathbf{a},\tmmathbf{y})$ as an approximation only functions during inference by producing samples to compute the stochastic gradients, while $p_{\theta}(\tmmathbf{h} |\tmmathbf{q})$ is the generative distribution that generates the samples for predicting the question-answer relatedness $\tmmathbf{y}$.

Based on the samples $\tmmathbf{h} \sim q_{\phi} ( \tmmathbf{h} | \tmmathbf{q}, \tmmathbf{a}, \tmmathbf{y})$, we use SGVB to optimise the lower bound (Eq.\ref{eq:lb_nasm}). 
The model parameters $\theta$ and the inference network parameters $\phi$ are updated jointly using their stochastic gradients. 
In this case, similar to the NVDM, the Gaussian KL divergence $D_{\tmop{KL}} [q_{\phi} ( \tmmathbf{h} | \tmmathbf{q}, \tmmathbf{a}, \tmmathbf{y}))\|p_{\theta} ( \tmmathbf{h} | \tmmathbf{q}) ]$ can be analytically computed during training process. 
\begin{table*}[!tb] 
\subfloat[Perplexity on test dataset.]{
  \centering
    \small
    \addtolength{\tabcolsep}{-2.0pt}
    \renewcommand{\arraystretch}{1.1}
\begin{tabular}{l|r|r|r}
\toprule[1.2pt]
	Model & Dim & 20News & RCV1 \\
	\hline
	LDA & 50 & 1091 & 1437 \\
	LDA & 200 & 1058 & 1142 \\
	RSM & 50 & 953 & 988 \\
	docNADE & 50 & 896 & 742 \\
	\hline
	SBN & 50 & 909 & 784 \\
	fDARN & 50 & 917 & 724 \\
	fDARN & 200 & ---- & 598 \\
	\hline
	NVDM & 50 & \textbf{836} & 563 \\
	NVDM & 200 & 852 & \textbf{550} \\
\bottomrule[1.2pt]
\end{tabular}
\label{tb:ppx}
}
\hfill
\subfloat[The five nearest words in the semantic space.]{
\centering
\addtolength{\tabcolsep}{-1.0pt}
\small
\begin{tabular}{l|c|c|c|c|c|c}
\toprule[1.2pt]
Word & \textbf{weapons} 	& \textbf{medical} 	& \textbf{companies}	& \textbf{define}		& \textbf{israel} 	& \textbf{book} \\
 \hline
	& guns		& medicine	& expensive	& defined	& israeli	& books		 \\
	& weapon	 	& health	 	& industry	& definition	& arab		& reference	 \\
NVDM	& gun		& treatment 	& company	& printf		& arabs		& guide		 \\
	& militia	& disease	& market		& int		& lebanon	& writing	 \\
	& armed		& patients	& buy		& sufficient	& lebanese	& pages		 \\
\hline
& weapon		& treatment	& demand		& defined	& israeli		& reading	 \\
	& shooting	&  medecine 	& commercial	& definition	&  israelis		&  read		 \\
NADE		& firearms	&  patients 	& agency 	& refer		&  arab			&  books		 \\
	& assault	&  process 	& company 	& make		&  palestinian	&  relevent	 \\
	& armed		&  studies 	& credit 	& examples	&  arabs			&  collection	 \\
\bottomrule[1.2pt]
\end{tabular}
\label{tb:nw}   
}
\caption{For the experimental results in (a), LDA \citep{blei2003latent} is a traditional topic model that models documents by mixtures of topics, RSM \citep{hinton2009replicated} is an undirected topic model implemented by restricted Boltzmann machines, and docNADE \citep{larochelle2012neural} is a neural topic model based on autoregressive assumption. The models based on Sigmoid Belief Networks (SBN) and Deep AutoRegressive Neural Network (DARN) structures are implemented by \citet{mnih2014neural}, which employs an MLP to build a Monte Carlo control variate estimator for stochastic estimation. }
\end{table*}

\section{Experiments}
\subsection{Dataset \& Setup for Document Modelling}
We experiment with NVDM on two standard news corpora: the \textit{20NewsGroups}\footnote{http://qwone.com/~jason/20Newsgroups} and the Reuters \textit{RCV1-v2}\footnote{http://trec.nist.gov/data/reuters/reuters.html}. 
The former is a collection of newsgroup documents, consisting of 11,314 training and 7,531 test articles. 
The latter is a large collection from Reuters newswire stories with 794,414 training and 10,000 test cases. 
The vocabulary size of these two datasets are set as 2,000 and 10,000.

To make a direct comparison with the prior work we follow the same preprocessing procedure and setup as \citet{hinton2009replicated}, \citet{larochelle2012neural}, \citet{Srivastava2013}, and \citet{mnih2014neural}.
We train NVDM models with 50 and 200 dimensional document representations respectively. 
For the inference network, we use an MLP (Eq. \ref{eq:mlp}) with 2 layers and 500 dimension rectifier linear units, which converts document representations into embeddings.
During training we carry out stochastic estimation by taking one sample for estimating the stochastic gradients, while in prediction we use 20 samples for predicting document perplexity. 
The model is trained by Adam \citep{DBLP:journals/corr/KingmaB14} and tuned by hold-out validation perplexity. We alternately optimise the generative model and the inference network by fixing the parameters of one while updating the parameters of the other.

\subsection{Experiments on Document Modelling}
Table \ref{tb:ppx} presents the test document perplexity. The first column lists the models, and the second column shows the dimension of latent variables used in the experiments. 
The final two columns present the perplexity achieved by each topic model on the \textit{20NewsGroups} and \textit{RCV1-v2} datasets.
In document modelling, perplexity is computed by $exp(-\frac{1}{D}\sum_n^{N_{d}} \frac{1}{N_d} \log p(\tmmathbf{X}_d))$, where $D$ is the number of documents, $N_d$ represents the length of the $d$th document and $\log p(\tmmathbf{X}) = \log \int p(\tmmathbf{X}|\tmmathbf{h}) p(\tmmathbf{h})d\tmmathbf{h}$ is the log probability of the words in the document. 
Since $\log p(\tmmathbf{X})$ is intractable in the NVDM, we use the variational lower bound (which is an upper bound on perplexity) to compute the perplexity following \citet{mnih2014neural}.

While all the baseline models listed in Table \ref{tb:ppx} apply discrete latent variables, here NVDM employs a continuous stochastic document representation. The experimental results indicate that NVDM achieves the best performance on both datasets. 
For the experiments on \textit{RCV1-v2} dataset, the NVDM with latent variable of 50 dimension performs even better than the fDARN with 200 dimension. 
It demonstrates that our document model with continuous latent variables has higher expressiveness and better generalisation ability. Table \ref{tb:nw} compares the 5 nearest words selected according to the semantic vector learned from NVDM and docNADE.
\begin{table}[tb]
\addtolength{\tabcolsep}{-3.5pt}
  \small
  \centering
\begin{tabular}{c|c|c|c|c}
\toprule[1.2pt]
	\textit{\textbf{Space}}&		\textit{\textbf{Religion}}&	\textit{\textbf{Encryption}}&		\textit{\textbf{Sport}}&		\textit{\textbf{Policy}}	\\
\hline
	orbit&		muslims& 		rsa&			goals&		bush \\
	lunar&		worship& 		cryptography&	pts&		resources\\
	solar&		belief&  		crypto&		teams&		charles\\
	shuttle& 	genocide&		keys&		league&		austin\\
	moon&		jews&			pgp& 		team	& 		bill\\
	launch&		islam&			license& 	players&		resolution\\
	fuel&		christianity&	secure&		nhl&			mr\\
	nasa&   		atheists&		key& 		stats&		misc\\
	satellite&	muslim&			escrow& 		min& 		piece\\
	japanese&	religious&		trust&		buf& 		marc\\
\bottomrule[1.2pt]
\end{tabular} 
\caption{The topics learned by NVDM on 20News.}
\label{tb:topics}
\end{table}

In addition to the perplexities, we also qualitatively evaluate the semantic information learned by NVDM on the \textit{20NewsGroups} dataset with latent variables of 50 dimension. 
We assume each dimension in the latent space represents a topic that corresponds to a specific semantic meaning.
Table \ref{tb:topics} presents 5 randomly selected topics with 10 words that have the strongest positive connection with the topic.
Based on the words in each column, we can deduce their corresponding topics as: \textit{Space, Religion, Encryption, Sport} and \textit{Policy}. Although the model does not impose independent interpretability on the latent representation dimensions, we still see that the NVDM learns locally interpretable structure.
\begin{table*}[t]
\begin{minipage}[b]{0.48\textwidth} 
  \begin{minipage}[b]{1\textwidth}        
  \centering
    \small
    \addtolength{\tabcolsep}{-2.0pt}
    \renewcommand{\arraystretch}{1.1}
	\begin{tabular}{@{}l|lrrr@{}}
			\toprule
			Source& Set &  Questions &  QA Pairs  & Judgement\\
			\hline
			&Train  & 1,229 & 53,417  & automatic \\
			QASent&Dev		   & 82	  & 1,148   & manual \\
			&Test	   & 100  & 1,517   & manual \\
			\hline
			&Train  & 2,118 & 20,360  & manual \\
			WikiQA&Dev		   & 296	  & 2,733   & manual \\
			&Test	   & 633  & 6,165   & manual \\
			\bottomrule
	\end{tabular}
	\caption{  Statistics of \textit{QASent} and \textit{WikiQA}. Judgement denotes whether correctness was determined automatically or by human annotators.
	}\label{dataset}
  \end{minipage}
   \begin{minipage}[b]{1\textwidth}   
  \centering
	\includegraphics[width=2.8in]{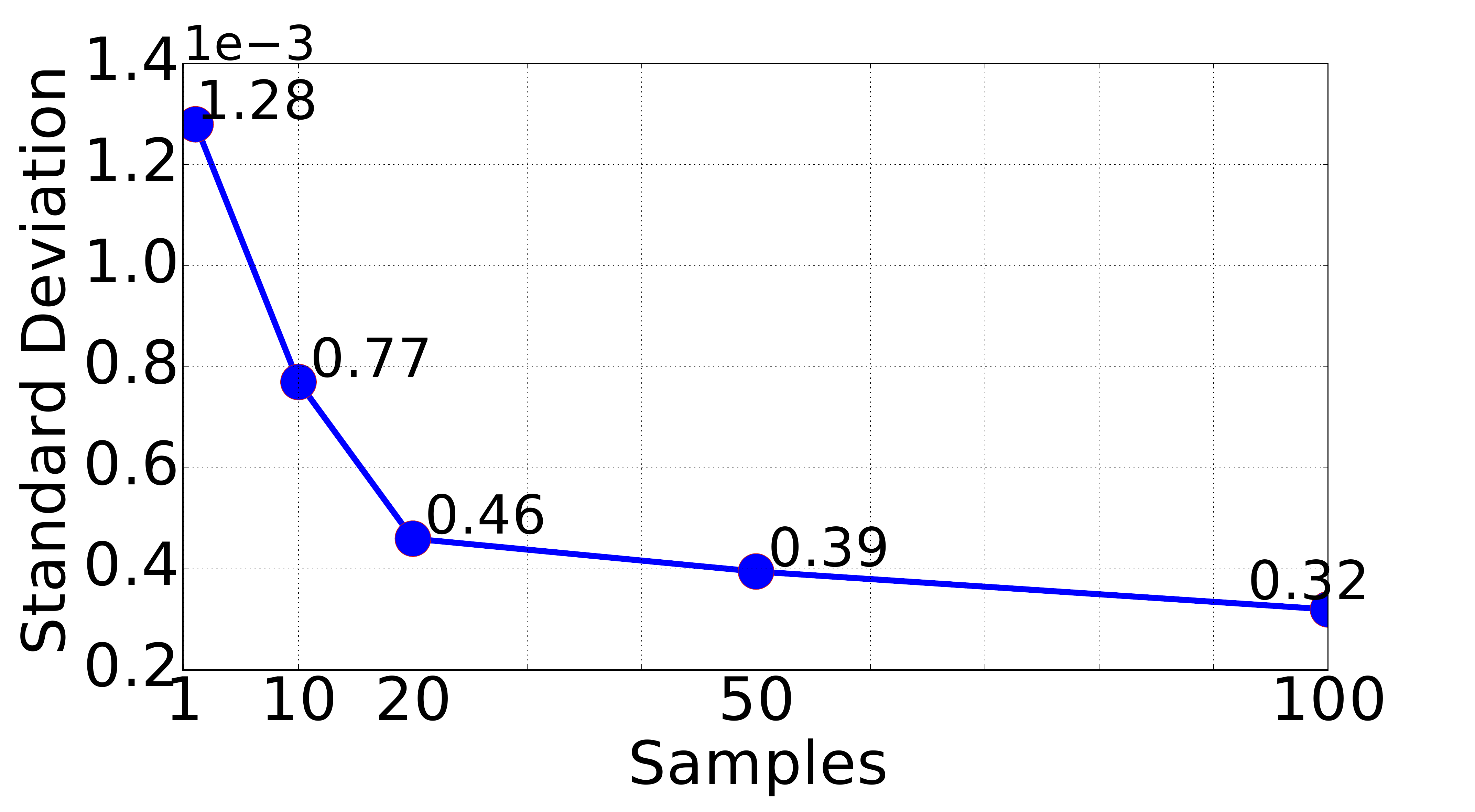}
	\captionof{figure}{The standard deviations of MAP scores computed by running 10 NASM models on WikiQA with different numbers of samples. }\label{fig:var}
  \end{minipage}
\end{minipage}
\hfill
\begin{minipage}[b]{0.48\textwidth} 
\centering
\small
\addtolength{\tabcolsep}{-1.0pt}
\renewcommand{\arraystretch}{0.9}
  \begin{tabular}{@{}llrrrrr@{}}
  	\toprule
  	&\multirow{2}{1cm}{Model}&\multicolumn{2}{c}{QASent} && \multicolumn{2}{c}{WikiQA}\\
  	\cmidrule{3-4}\cmidrule{6-7}
  	& & MAP & MRR & &MAP& MRR\\
  	\midrule
  	\multicolumn{2}{@{}l}{\textbf{Published Models}} && &&& \\
  	& PV							&0.5213 &0.6023 & & 0.5110 & 0.5160\\
  	& Bigram-CNN 			& 0.5693	&  0.6613 &&0.6190 &0.6281\\
  	& Deep CNN 			& 0.5719	& 0.6621 & &--- &---\\
  	& PV + Cnt &  0.6762 & 0.7514& &0.5976 & 0.6058 \\
  	& WA					 & 0.7063 & 0.7740 &&--- &---\\
  	& LCLR 	&  0.7092	& 0.7700 && 0.5993 & 0.6068\\
  	& Bigram-CNN + Cnt 			& 0.7113	& 0.7846 && 0.6520 & 0.6652\\
  	& Deep CNN + Cnt 			& 0.7186	& 0.7826 &&--- &---\\
  	\midrule
  	\multicolumn{2}{@{}l}{\textbf{Our Models}} &&  && &\\
  	& LSTM & 0.6436 & 0.7235 && 0.6552	& 0.6747 \\
  	& LSTM + Att & 0.6451 & 0.7316 &&  0.6639 &  0.6828 \\
  	& NASM & \bf{0.6501} & \bf{0.7324} && \bf{0.6705} & \bfseries{0.6914} \\
  	& LSTM + Cnt & 0.7228	& 0.7986 && 0.6820 & 0.6988\\
  	& LSTM + Att + Cnt &  0.7289 & 0.8072 && 0.6855 & 0.7041 \\ 
  	& NASM + Cnt & 	\bf{0.7339} & \bf{0.8117} &&  \bfseries{0.6886}& \bfseries{0.7069} \\
  	\bottomrule
  \end{tabular}
  \label{oldqa-result}
  \caption{Results of our models (LSTM, LSTM + Att, NASM) in comparison with other state of the art models on the \textit{QASent} and \textit{WikiQA} dataset. PV is the paragraph vector \citep{DBLP:conf/icml/LeM14}. Bigram-CNN is the simple convolutional model reported in \citep{Yu:2014}. Deep CNN is the deep convolutional model from \citep{severyn2015disi}. WA is a model based on word alignment \citep{wang2015faq}. LCLR is the SVM-based classifier trained using a set of features.  Model + Cnt means that the result is obtained from a combination of a lexical overlap feature and the output from the distributional model. }
 \label{result_table}
\end{minipage}
\end{table*}
\subsection{Dataset \& Setup for Answer Sentence Selection}

We experiment on two answer selection datasets, the \textit{QASent} and the \textit{WikiQA} datasets. 
\textit{QASent} \citep{wang2007jeopardy} is created from the TREC QA track, and the \textit{WikiQA} \citep{yang-yih-meek:2015:EMNLP} is constructed from Wikipedia, which is less noisy and less biased towards lexical overlap\footnote{\citet{yang-yih-meek:2015:EMNLP} provide detailed explanation of the differences between the two datasets.}. Table \ref{dataset} summarises the statistics of the two datasets.

In order to investigate the effectiveness of our NASM model we also implemented two strong baseline models --- a vanilla LSTM model (LSTM) and an LSTM model with a deterministic attention mechanism (LSTM+Att). 
The former directly applies the QA matching function (Eq. \ref{qa_match}) on the independent question and answer representations which are the last state outputs $\tmmathbf{s}_q (| \tmmathbf{q} |)$ and $\tmmathbf{s}_a (|\tmmathbf{a} |)$ from the question and answer LSTM models. 
The latter adds an attention model to learn pair-specific representation for prediction on the basis of the vanilla LSTM. 
Moreover, LSTM+Att is the deterministic counterpart of NASM, which has the same neural network architecture as NASM. The only difference is that it replaces the stochastic units $\tmmathbf{h}$ with deterministic ones, and no inference network is required to carry out stochastic estimation.
Following previous work, for each of our models we also add a lexical overlap feature by combining a co-occurrence word count feature with the probability generated from the neural model. 
MAP and MRR are adopted as the evaluation metrics for this task.

To facilitate direct comparison with previous work we follow the same experimental setup as \citet{Yu:2014} and \citet{severyn2015disi}. 
The word embeddings $(K = 50)$  are obtained by running the \texttt{word2vec} tool \citep{DBLP:conf/nips/MikolovSCCD13} on the English Wikipedia dump and the \textit{AQUAINT}\footnote{\url{https://catalog.ldc.upenn.edu/LDC2002T31}} corpus. 
We use LSTMs with $3$ layers and $50$ hidden units, and apply $40 \%$ dropout after the embedding layer. 
For the construction of the inference network, we use an MLP (Eq. \ref{eq:mlp2}) with 2 layers and tanh units of 50 dimension, and an MLP (Eq. \ref{eq:jointMLP}) with 2 layers and tanh units of 150 dimension for modelling the joint representation.
During training we carry out stochastic estimation by taking one sample for computing the gradients, while in prediction we use 20 samples to calculate the expectation of the lower bound. 
Figure \ref{fig:var} presents the standard deviation of NASM's MAP scores while using different numbers of samples. 
Considering the trade-off between computational cost and variance, we chose 20 samples for prediction in all the experiments. 
The models are trained using Adam \citep{DBLP:journals/corr/KingmaB14}, with hyperparameters selected by optimising the MAP score on the development set.

\subsection{Experiments on Answer Sentence Selection}
\label{result}
\begin{figure*}[!htb]
\begin{minipage}[t]{0.53\linewidth}
  \centering
	\includegraphics[width=4.0in]{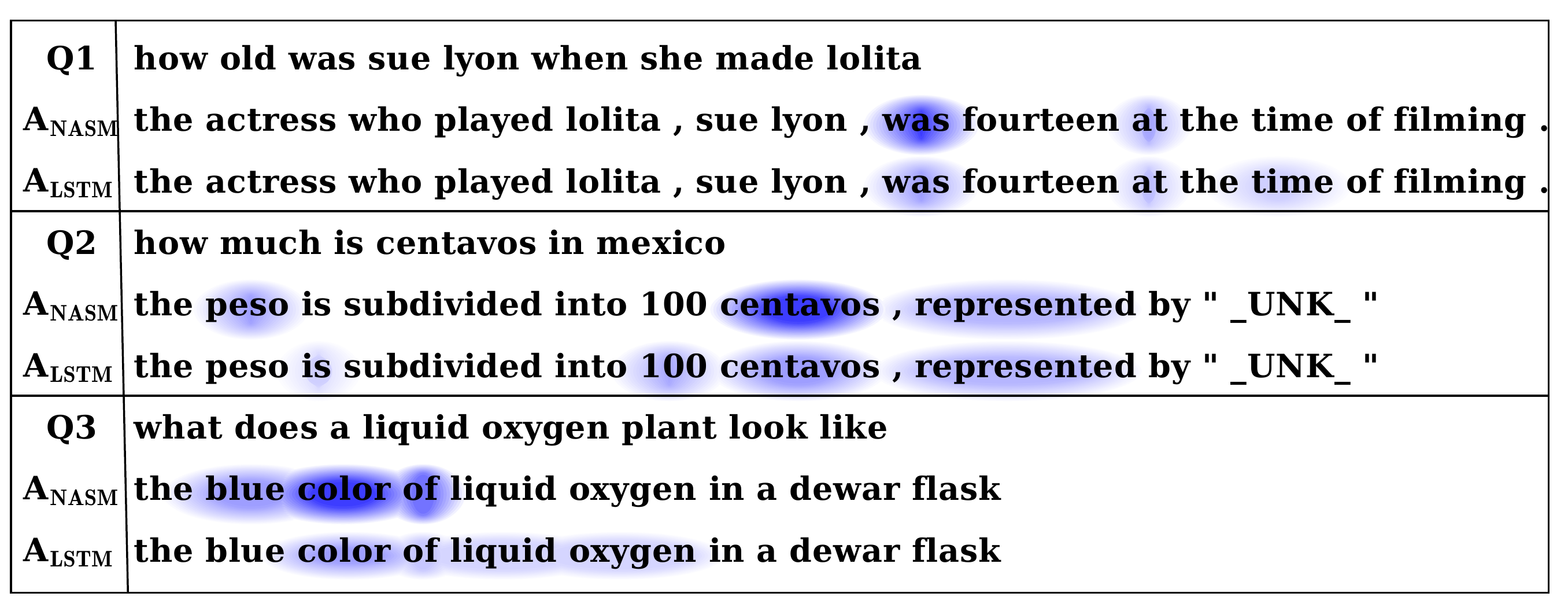}
	\caption{A visualisation of attention scores on answer sentences.}
	\label{fig:att}
\end{minipage}
\hfill
\begin{minipage}[t]{0.5\linewidth}
  \centering
	\includegraphics[width=2.2in]{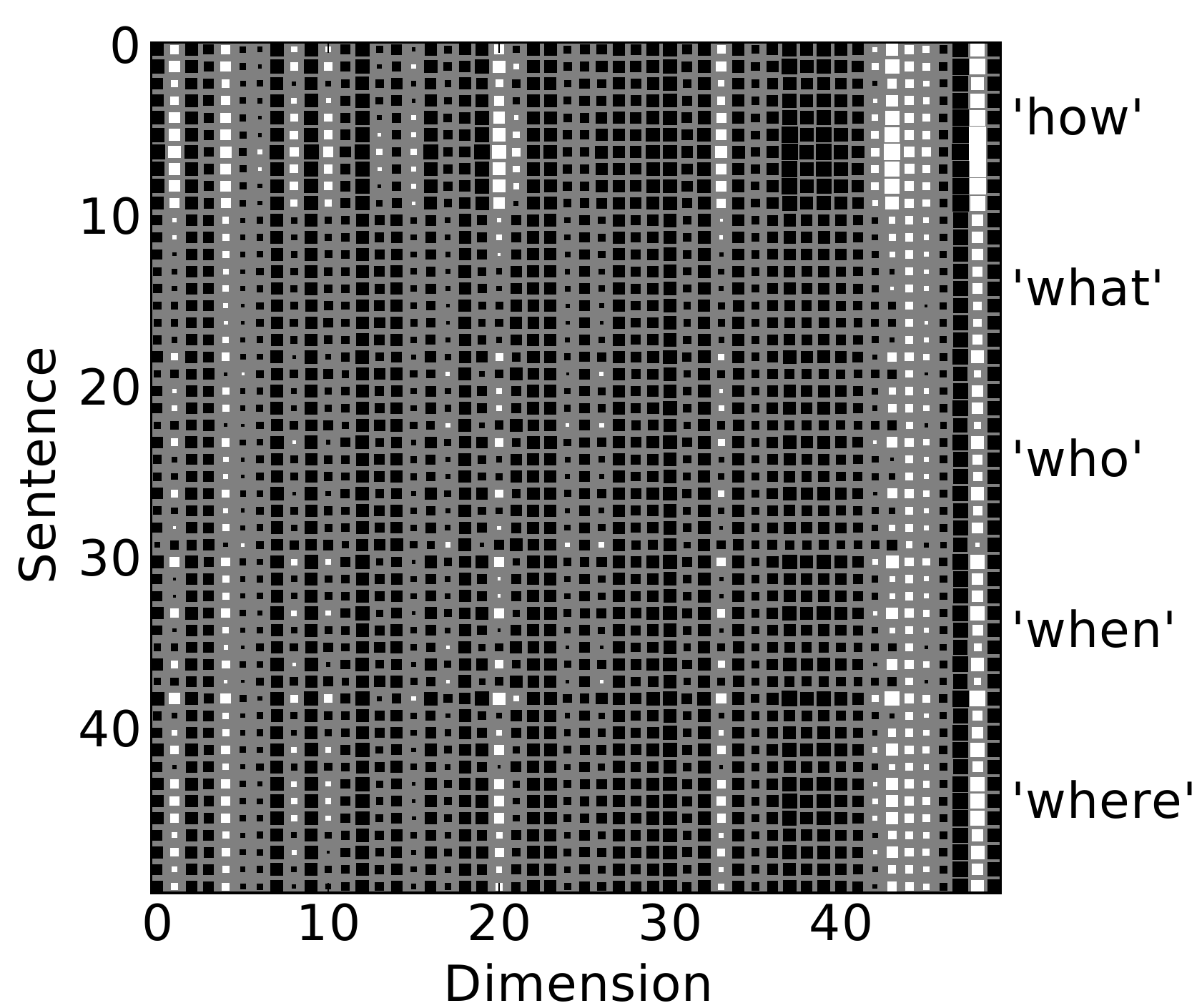}
	\caption{Hinton diagrams of the log standard deviations.}
	\label{fig:hinton}
\end{minipage}
\end{figure*}

Table \ref{result_table} compares the results of our models with current state-of-the-art models on both answer selection datasets. 
On the \textit{QASent} dataset, our vanilla LSTM model outperforms the deep CNN \footnote{As stated in \citep{yih2013question} that the evaluation scripts used by previous work are noisy --- 4 out of 72 questions in the test set are treated answered incorrectly. 
This makes the MAP and MRR scores $\sim4\%$ lower than the {\it true} scores. 
Since \citet{severyn2015disi} and \citet{wang2015faq} use a cleaned-up evaluation scripts, we apply the original noisy scripts to re-evaluate their outputs in order to make the results directly comparable with previous work. } model by approximately $7\%$ on MAP and $6\%$ on MRR. 
The LSTM+Att performs slightly better than the vanilla LSTM model, and our NASM improves the results further. 
Since the \textit{QASent} dataset is biased towards lexical overlapping features, after combining with a co-occurrence word count feature, our best model NASM outperforms all the previous models, including both neural network based models and classifiers with a set of hand-crafted features (e.g. LCLR). 
Similarly, on the \textit{WikiQA} dataset, all of our models outperform the previous distributional models by a large margin. 
By including a word count feature, our models improve further and achieve the state-of-the-art. 
Notably, on both datasets, our two LSTM-based models have set strong baselines and NASM works even better, which demonstrates the effectiveness of introducing stochastic units to model question semantics in this answer sentence selection task.

In Figure \ref{fig:att}, we compare the effectiveness of the latent attention mechanism (NASM) and its deterministic counterpart (LSTM+Att) by visualising the attention scores on the answer sentences. 
For most of the negative answer sentences, neither of the two attention models can attend to reasonable words that are beneficial for predicting relatedness. 
But for the correct answer sentences, such as the ones in Figure \ref{fig:att}, both attention models are able to capture crucial information by attending to different parts of the sentence based on the question semantics. 
Interestingly, compared to the deterministic counterpart LSTM+Att, our NASM assigns higher attention scores on the prominent words that are relevant to the question, which forms a more peaked distribution and in turn helps the model achieve better performance. 

In order to have an intuitive observation on the latent distributions, we present Hinton diagrams of their log standard deviation parameters (Figure \ref{fig:hinton}). In a Hinton diagram, the size of a square is proportional to a value's magnitude, and the colour (black/white) indicates its sign (positive/negative). In this case, we visualise the parameters of 50 conditional distributions $p_{\theta}(\tmmathbf{h}|\tmmathbf{q})$ with the questions selected from 5 different groups, which start with `how', `what', `who', `when' and `where'. All the log standard deviations are initialised as zero before training.  According to Figure \ref{fig:hinton}, we can see that the questions starting with `how' have more white areas, which indicates higher variances or more uncertainties are in these dimensions. By contrast, the questions starting with `what' have black squares in almost every dimension. Intuitively, it is more difficult to understand and answer the questions starting with `how' than the others, while the `what' questions commonly have explicit words indicating the possible answers. To validate this, we compute the stratified MAP scores based on different question type. The MAP of 'how' questions is 0.524 which is the lowest among the five groups. Hence empirically,  'how' questions are harder to 'understand and answer'.

\section{Discussion}
As shown in the experiments, neural variational inference brings consistent improvements on the performance of both NLP tasks. The basic intuition is that the latent distributions grant the ability to sum over all the possibilities in terms of semantics. From the perspective of optimisation, one of the most important reasons is that Bayesian learning guards against overfitting. 

According to Eq. \ref{eq:lb} in NVDM, since we adopt $p(\tmmathbf{h})$ as a standard Gaussian prior, the KL divergence term $D_{\tmop{KL}} [q_{\phi} ( \tmmathbf{h} | \tmmathbf{X})\|p (\tmmathbf{h} )]$ can be analytically computed as $\frac{1}{2}(K-\| \tmmathbf{\mu} \|^2 - \| \tmmathbf{\sigma} \|^2 + \log | \tmop{diag} ( \tmmathbf{\sigma}^2) |)$. It is not difficult to find that it actually acts as L2 regulariser when we update the $\tmmathbf{\mu}$. 
Similarly, in NASM (Eq. \ref{eq:lb_nasm}), we also have the KL divergence term $D_{\tmop{KL}} [q_{\phi} ( \tmmathbf{h} | \tmmathbf{q}, \tmmathbf{a}, \tmmathbf{y}))\|p_{\theta} ( \tmmathbf{h} | \tmmathbf{q}) ]$. 
Different from NVDM, it attempts to minimise the distance between $q_{\phi} ( \tmmathbf{h} | \tmmathbf{q}, \tmmathbf{a}, \tmmathbf{y}))$ and $p_{\theta} ( \tmmathbf{h} | \tmmathbf{q})$ that are both conditional distributions. 
Because $p_{\theta} ( \tmmathbf{h} | \tmmathbf{q})$ as well as $q_{\phi} ( \tmmathbf{h} | \tmmathbf{q}, \tmmathbf{a}, \tmmathbf{y}))$ are learned during training, the two distributions are mutually restrained while being updated. 
Therefore, NVDM simply penalises the large $\tmmathbf{\mu}$ and encourages $q_{\phi} ( \tmmathbf{h} | \tmmathbf{X})$ to approach the prior $p(\tmmathbf{h})$ for every document $\tmmathbf{X}$, 
but in NASM, $p_{\theta} ( \tmmathbf{h} | \tmmathbf{q})$ acts like a moving baseline distribution which regularises the update of $q_{\phi} ( \tmmathbf{h} | \tmmathbf{q}, \tmmathbf{a}, \tmmathbf{y}))$ for every different conditions. 
In practice, we carry out early stopping by observing the prediction performance on development dataset for the question answer selection task. Using the same learning rate and neural network structure, LSTM+Att reaches optimal performance and starts to overfit on training dataset generally at the $20$th iteration, while NASM starts to overfit around the $35$th iteration. 

More interestingly, in the question answer selection experiments, NASM learns more peaked attention scores than its deterministic counterpart LSTM+Att. 
For the update process of LSTM+Att, we find there exists a relatively big variance in the gradients w.r.t. question semantics (LSTM+Att applies deterministic $\tmmathbf{s}_q(|\tmmathbf{q}|)$ while NASM applies stochastic $\tmmathbf{h}$). This is because the training dataset is small and contains many negative answer sentences that brings no benefit but noise to the learning of the attention model.
In contrast, for the update process of NASM, we observe more stable gradients w.r.t. the parameters of latent distributions. 
The optimisation of the lower bound on one hand maximises the conditional log-likelihood (that the deterministic counterpart cares about) and on the other hand minimises the KL-divergence (that regularises the gradients). 
Hence, each update of the lower bound actually keeps the gradients w.r.t. $\tmmathbf{\mu}$ from swinging heavily.
Besides, since the values of $\tmmathbf{\sigma}$ are not very significant in this case, the distribution of attention scores mainly depends on $\tmmathbf{\mu}$. Therefore, the learning of the attention model benefits from the regularisation as well, and it explains the fact that NASM learns more peaked attention scores which in turn helps achieve a better prediction performance.

Since the computations of NVDM and NASM can be parallelised on GPU and only one sample is required during training process, it is very efficient to carry out the neural variational inference. Moreover, for both NVDM and NASM, all the parameters are updated by back-propagation. Thus, the increased computation time for the stochastic units only comes from the added parameters of the inference network.

\section{Related Work}
Training an inference network to approximate the variational distribution was first proposed in the context of Helmholtz machines \citep{hinton1994autoencoders,hinton1995wake,dayan1996varieties}, but applications of these directed generative models come up against the problem of establishing low variance gradient estimators. 
Recent advances in neural variational inference mitigate this problem by reparameterising the continuous random variables \citep{rezende2014stochastic,kingma2013auto}, using control variates \citep{mnih2014neural} or approximating the posterior with importance sampling \citep{bornschein2014reweighted}.  
The instantiations of these ideas \citep{gregor2015draw,kingma2014semi,ba2015learning} have demonstrated strong performance on the tasks of image processing. The recent variants of generative auto-encoder 
\citep{louizos2015variational,DBLP:journals/corr/MakhzaniSJG15} are also very competitive. 
\citet{tang2013learning} applies the similar idea of introducing stochastic units for expression classification, but its inference is carried out by Monte Carlo EM algorithm with the reliance on importance sampling, which is less efficient and lack of scalability.

Another class of neural generative models make use of the autoregressive assumption \citep{larochelle2011neural,uria2014deep,germain2015made,
gregor2013deep}. 
Applications of these models on document modelling achieve significant improvements on generating documents, compared to conventional probabilistic topic models \citep{hofmann1999probabilistic,blei2003latent} and also the RBMs \citep{hinton2009replicated,Srivastava2013}. 
While these models that use binary semantic vectors, our NVDM employs dense continuous document representations which are both expressive and easy to train. 
The semantic word vector model \citep{maas2011learning} also employs a continuous semantic vector to generate words, but the model is trained by MAP inference which does not permit the calculation of the posterior distribution.
A very similar idea to NVDM is \citet{DBLP:journals/corr/BowmanVVDJB15}, which employs VAE to generate sentences from a continuous space.

Apart from the work mentioned above, there is other interesting work on question answering with deep neural networks. One of the popular streams is mapping factoid questions with answer triples in the knowledge base \citep{Bordes:2014:EMNLP,DBLP:journals/corr/BordesWU14,DBLP:conf/acl/YihHM14}. Moreover, \citet{DBLP:journals/corr/WestonCB14,sukhbaatar2015end,DBLP:journals/corr/KumarISBEPOGS15} further exploit memory networks, where long-term memories act as dynamic knowledge bases. Another attention-based model \citep{DBLP:journals/corr/HermannKGEKSB15} applies the attentive network to help read and comprehend for long articles.

\section{Conclusion}
This paper introduced a deep neural variational inference framework for generative models of text. We experimented on two diverse tasks, document modelling and question answer selection tasks to demonstrate the effectiveness of this framework, where in both cases our models achieve state of the art performance. 
Apart from the promising results, our model also has the advantages of (1) simple, expressive, and efficient when training with the SGVB algorithm; (2) suitable for both unsupervised and supervised learning tasks; and (3) capable of generalising to incorporate any type of neural network.

\newpage
\bibliography{iclr2016_conference}

\begin{thebibliography}{47}
\providecommand{\natexlab}[1]{#1}
\providecommand{\url}[1]{\texttt{#1}}
\expandafter\ifx\csname urlstyle\endcsname\relax
  \providecommand{\doi}[1]{doi: #1}\else
  \providecommand{\doi}{doi: \begingroup \urlstyle{rm}\Url}\fi

\bibitem[Andrieu et~al.(2003)Andrieu, De~Freitas, Doucet, and
  Jordan]{andrieu2003introduction}
Andrieu, Christophe, De~Freitas, Nando, Doucet, Arnaud, and Jordan, Michael~I.
\newblock An introduction to mcmc for machine learning.
\newblock \emph{Machine learning}, 50\penalty0 (1-2):\penalty0 5--43, 2003.

\bibitem[Attias(2000)]{attias2000variational}
Attias, Hagai.
\newblock A variational bayesian framework for graphical models.
\newblock In \emph{Proceedings of NIPS}, 2000.

\bibitem[Ba et~al.(2015)Ba, Grosse, Salakhutdinov, and Frey]{ba2015learning}
Ba, Jimmy, Grosse, Roger, Salakhutdinov, Ruslan, and Frey, Brendan.
\newblock Learning wake-sleep recurrent attention models.
\newblock In \emph{Proceedings of NIPS}, 2015.

\bibitem[Bahdanau et~al.(2015)Bahdanau, Cho, and Bengio]{bahdanau2014neural}
Bahdanau, Dzmitry, Cho, Kyunghyun, and Bengio, Yoshua.
\newblock Neural machine translation by jointly learning to align and
  translate.
\newblock In \emph{Proceedings of ICLR}, 2015.

\bibitem[Beal(2003)]{beal2003variational}
Beal, Matthew~James.
\newblock \emph{Variational algorithms for approximate Bayesian inference}.
\newblock University of London, 2003.

\bibitem[Blei et~al.(2003)Blei, Ng, and Jordan]{blei2003latent}
Blei, David~M, Ng, Andrew~Y, and Jordan, Michael~I.
\newblock Latent dirichlet allocation.
\newblock \emph{The Journal of Machine Learning Research}, 3:\penalty0
  993--1022, 2003.

\bibitem[Bordes et~al.(2014{\natexlab{a}})Bordes, Chopra, and
  Weston]{Bordes:2014:EMNLP}
Bordes, Antoine, Chopra, Sumit, and Weston, Jason.
\newblock Question answering with subgraph embeddings.
\newblock In \emph{Proceedings of EMNLP}, 2014{\natexlab{a}}.

\bibitem[Bordes et~al.(2014{\natexlab{b}})Bordes, Weston, and
  Usunier]{DBLP:journals/corr/BordesWU14}
Bordes, Antoine, Weston, Jason, and Usunier, Nicolas.
\newblock Open question answering with weakly supervised embedding models.
\newblock In \emph{Proceedings of ECML}, 2014{\natexlab{b}}.

\bibitem[Bornschein \& Bengio(2015)Bornschein and
  Bengio]{bornschein2014reweighted}
Bornschein, J{\"o}rg and Bengio, Yoshua.
\newblock Reweighted wake-sleep.
\newblock In \emph{Proceedings of ICLR}, 2015.

\bibitem[Bowman et~al.(2015)Bowman, Vilnis, Vinyals, Dai, J{\'{o}}zefowicz, and
  Bengio]{DBLP:journals/corr/BowmanVVDJB15}
Bowman, Samuel~R., Vilnis, Luke, Vinyals, Oriol, Dai, Andrew~M.,
  J{\'{o}}zefowicz, Rafal, and Bengio, Samy.
\newblock Generating sentences from a continuous space.
\newblock \emph{CoRR}, abs/1511.06349, 2015.
\newblock URL \url{http://arxiv.org/abs/1511.06349}.

\bibitem[Dayan \& Hinton(1996)Dayan and Hinton]{dayan1996varieties}
Dayan, Peter and Hinton, Geoffrey~E.
\newblock Varieties of helmholtz machine.
\newblock \emph{Neural Networks}, 9\penalty0 (8):\penalty0 1385--1403, 1996.

\bibitem[Germain et~al.(2015)Germain, Gregor, Murray, and
  Larochelle]{germain2015made}
Germain, Mathieu, Gregor, Karol, Murray, Iain, and Larochelle, Hugo.
\newblock Made: Masked autoencoder for distribution estimation.
\newblock In \emph{Proceedings of ICML}, 2015.

\bibitem[Gregor et~al.(2014)Gregor, Mnih, and Wierstra]{gregor2013deep}
Gregor, Karol, Mnih, Andriy, and Wierstra, Daan.
\newblock Deep autoregressive networks.
\newblock In \emph{Proceedings of ICML}, 2014.

\bibitem[Gregor et~al.(2015)Gregor, Danihelka, Graves, and
  Wierstra]{gregor2015draw}
Gregor, Karol, Danihelka, Ivo, Graves, Alex, and Wierstra, Daan.
\newblock Draw: A recurrent neural network for image generation.
\newblock In \emph{Proceedings of ICML}, 2015.

\bibitem[Hermann et~al.(2015)Hermann, Kocisk{\'{y}}, Grefenstette, Espeholt,
  Kay, Suleyman, and Blunsom]{DBLP:journals/corr/HermannKGEKSB15}
Hermann, Karl~Moritz, Kocisk{\'{y}}, Tom{\'{a}}s, Grefenstette, Edward,
  Espeholt, Lasse, Kay, Will, Suleyman, Mustafa, and Blunsom, Phil.
\newblock Teaching machines to read and comprehend.
\newblock In \emph{Proceedings of NIPS}, 2015.

\bibitem[Hinton \& Salakhutdinov(2009)Hinton and
  Salakhutdinov]{hinton2009replicated}
Hinton, Geoffrey~E and Salakhutdinov, Ruslan.
\newblock Replicated softmax: an undirected topic model.
\newblock In \emph{Proceedings of NIPS}, 2009.

\bibitem[Hinton \& Zemel(1994)Hinton and Zemel]{hinton1994autoencoders}
Hinton, Geoffrey~E and Zemel, Richard~S.
\newblock Autoencoders, minimum description length, and helmholtz free energy.
\newblock In \emph{Proceedings of NIPS}, 1994.

\bibitem[Hinton et~al.(1995)Hinton, Dayan, Frey, and Neal]{hinton1995wake}
Hinton, Geoffrey~E, Dayan, Peter, Frey, Brendan~J, and Neal, Radford~M.
\newblock The wake-sleep algorithm for unsupervised neural networks.
\newblock \emph{Science}, 268\penalty0 (5214):\penalty0 1158--1161, 1995.

\bibitem[Hochreiter \& Schmidhuber(1997)Hochreiter and
  Schmidhuber]{hochreiter1997long}
Hochreiter, Sepp and Schmidhuber, J{\"u}rgen.
\newblock Long short-term memory.
\newblock \emph{Neural computation}, 9\penalty0 (8):\penalty0 1735--1780, 1997.

\bibitem[Hofmann(1999)]{hofmann1999probabilistic}
Hofmann, Thomas.
\newblock Probabilistic latent semantic indexing.
\newblock In \emph{Proceedings of SIGIR}, 1999.

\bibitem[Jordan et~al.(1999)Jordan, Ghahramani, Jaakkola, and
  Saul]{jordan1999introduction}
Jordan, Michael~I, Ghahramani, Zoubin, Jaakkola, Tommi~S, and Saul, Lawrence~K.
\newblock An introduction to variational methods for graphical models.
\newblock \emph{Machine learning}, 37\penalty0 (2):\penalty0 183--233, 1999.

\bibitem[Kingma \& Ba(2015)Kingma and Ba]{DBLP:journals/corr/KingmaB14}
Kingma, Diederik~P. and Ba, Jimmy.
\newblock Adam: {A} method for stochastic optimization.
\newblock In \emph{Proceedings of ICLR}, 2015.

\bibitem[Kingma \& Welling(2014)Kingma and Welling]{kingma2013auto}
Kingma, Diederik~P and Welling, Max.
\newblock Auto-encoding variational bayes.
\newblock In \emph{Proceedings of ICLR}, 2014.

\bibitem[Kingma et~al.(2014)Kingma, Mohamed, Rezende, and
  Welling]{kingma2014semi}
Kingma, Diederik~P, Mohamed, Shakir, Rezende, Danilo~Jimenez, and Welling, Max.
\newblock Semi-supervised learning with deep generative models.
\newblock In \emph{Proceedings of NIPS}, 2014.

\bibitem[Kumar et~al.(2015)Kumar, Irsoy, Su, Bradbury, English, Pierce,
  Ondruska, Gulrajani, and Socher]{DBLP:journals/corr/KumarISBEPOGS15}
Kumar, Ankit, Irsoy, Ozan, Su, Jonathan, Bradbury, James, English, Robert,
  Pierce, Brian, Ondruska, Peter, Gulrajani, Ishaan, and Socher, Richard.
\newblock Ask me anything: Dynamic memory networks for natural language
  processing.
\newblock \emph{arXiv preprint arXiv:1506.07285}, 2015.

\bibitem[Larochelle \& Lauly(2012)Larochelle and Lauly]{larochelle2012neural}
Larochelle, Hugo and Lauly, Stanislas.
\newblock A neural autoregressive topic model.
\newblock In \emph{Proceedings of NIPS}, 2012.

\bibitem[Larochelle \& Murray(2011)Larochelle and Murray]{larochelle2011neural}
Larochelle, Hugo and Murray, Iain.
\newblock The neural autoregressive distribution estimator.
\newblock In \emph{Proceedings of AISTATS}, 2011.

\bibitem[Le \& Mikolov(2014)Le and Mikolov]{DBLP:conf/icml/LeM14}
Le, Quoc~V. and Mikolov, Tomas.
\newblock Distributed representations of sentences and documents.
\newblock In \emph{Proceedings of ICML}, 2014.

\bibitem[Louizos et~al.(2015)Louizos, Swersky, Li, Welling, and
  Zemel]{louizos2015variational}
Louizos, Christos, Swersky, Kevin, Li, Yujia, Welling, Max, and Zemel, Richard.
\newblock The variational fair auto encoder.
\newblock \emph{arXiv preprint arXiv:1511.00830}, 2015.

\bibitem[Maas et~al.(2011)Maas, Daly, Pham, Huang, Ng, and
  Potts]{maas2011learning}
Maas, Andrew~L, Daly, Raymond~E, Pham, Peter~T, Huang, Dan, Ng, Andrew~Y, and
  Potts, Christopher.
\newblock Learning word vectors for sentiment analysis.
\newblock In \emph{Proceedings of ACL}, 2011.

\bibitem[Makhzani et~al.(2015)Makhzani, Shlens, Jaitly, and
  Goodfellow]{DBLP:journals/corr/MakhzaniSJG15}
Makhzani, Alireza, Shlens, Jonathon, Jaitly, Navdeep, and Goodfellow, Ian~J.
\newblock Adversarial autoencoders.
\newblock \emph{CoRR}, abs/1511.05644, 2015.
\newblock URL \url{http://arxiv.org/abs/1511.05644}.

\bibitem[Mikolov et~al.(2013)Mikolov, Sutskever, Chen, Corrado, and
  Dean]{DBLP:conf/nips/MikolovSCCD13}
Mikolov, Tomas, Sutskever, Ilya, Chen, Kai, Corrado, Gregory~S., and Dean,
  Jeffrey.
\newblock Distributed representations of words and phrases and their
  compositionality.
\newblock In \emph{Proceedings of NIPS}, 2013.

\bibitem[Mnih \& Gregor(2014)Mnih and Gregor]{mnih2014neural}
Mnih, Andriy and Gregor, Karol.
\newblock Neural variational inference and learning in belief networks.
\newblock In \emph{Proceedings of ICML}, 2014.

\bibitem[Neal(1993)]{neal1993probabilistic}
Neal, Radford~M.
\newblock Probabilistic inference using markov chain monte carlo methods.
\newblock \emph{Technical report: CRG-TR-93-1}, 1993.

\bibitem[Rezende et~al.(2014)Rezende, Mohamed, and
  Wierstra]{rezende2014stochastic}
Rezende, Danilo~J, Mohamed, Shakir, and Wierstra, Daan.
\newblock Stochastic backpropagation and approximate inference in deep
  generative models.
\newblock In \emph{Proceedings of ICML}, 2014.

\bibitem[Severyn(2015)]{severyn2015disi}
Severyn, Aliaksei.
\newblock \emph{Modelling input texts: from Tree Kernels to Deep Learning}.
\newblock PhD thesis, University of Trento, 2015.

\bibitem[Srivastava et~al.(2013)Srivastava, Salakhutdinov, and
  Hinton]{Srivastava2013}
Srivastava, Nitish, Salakhutdinov, RR, and Hinton, Geoffrey.
\newblock Modeling documents with deep boltzmann machines.
\newblock In \emph{Proceedings of UAI}, 2013.

\bibitem[Sukhbaatar et~al.(2015)Sukhbaatar, Szlam, Weston, and
  Fergus]{sukhbaatar2015end}
Sukhbaatar, Sainbayar, Szlam, Arthur, Weston, Jason, and Fergus, Rob.
\newblock End-to-end memory networks.
\newblock In \emph{Proceedings of NIPS}, 2015.

\bibitem[Tang \& Salakhutdinov(2013)Tang and Salakhutdinov]{tang2013learning}
Tang, Yichuan and Salakhutdinov, Ruslan~R.
\newblock Learning stochastic feedforward neural networks.
\newblock In \emph{Proceedings NIPS}, 2013.

\bibitem[Uria et~al.(2014)Uria, Murray, and Larochelle]{uria2014deep}
Uria, Benigno, Murray, Iain, and Larochelle, Hugo.
\newblock A deep and tractable density estimator.
\newblock In \emph{Proceedings of ICML}, 2014.

\bibitem[Wang et~al.(2007)Wang, Smith, and Mitamura]{wang2007jeopardy}
Wang, Mengqiu, Smith, Noah~A, and Mitamura, Teruko.
\newblock What is the jeopardy model? a quasi-synchronous grammar for qa.
\newblock In \emph{Proceedings of EMNLP-CoNLL}, 2007.

\bibitem[Wang \& Ittycheriah(2015)Wang and Ittycheriah]{wang2015faq}
Wang, Zhiguo and Ittycheriah, Abraham.
\newblock Faq-based question answering via word alignment.
\newblock \emph{arXiv preprint arXiv:1507.02628}, 2015.

\bibitem[Weston et~al.(2015)Weston, Chopra, and
  Bordes]{DBLP:journals/corr/WestonCB14}
Weston, Jason, Chopra, Sumit, and Bordes, Antoine.
\newblock Memory networks.
\newblock In \emph{Proceedings of ICLR}, 2015.

\bibitem[Yang et~al.(2015)Yang, Yih, and Meek]{yang-yih-meek:2015:EMNLP}
Yang, Yi, Yih, Wen-tau, and Meek, Christopher.
\newblock Wikiqa: A challenge dataset for open-domain question answering.
\newblock In \emph{Proceedings of EMNLP}, 2015.

\bibitem[Yih et~al.(2013)Yih, Chang, Meek, and Pastusiak]{yih2013question}
Yih, Wen-tau, Chang, Ming-Wei, Meek, Christopher, and Pastusiak, Andrzej.
\newblock Question answering using enhanced lexical semantic models.
\newblock In \emph{Proceedings of ACL}, 2013.

\bibitem[Yih et~al.(2014)Yih, He, and Meek]{DBLP:conf/acl/YihHM14}
Yih, Wen{-}tau, He, Xiaodong, and Meek, Christopher.
\newblock Semantic parsing for single-relation question answering.
\newblock In \emph{Proceedings of ACL}, 2014.

\bibitem[Yu et~al.(2014)Yu, Hermann, Blunsom, and Pulman]{Yu:2014}
Yu, Lei, Hermann, Karl~Moritz, Blunsom, Phil, and Pulman, Stephen.
\newblock {Deep Learning for Answer Sentence Selection}.
\newblock In \emph{{NIPS Deep Learning Workshop}}, 2014.

\end{thebibliography}
\bibliographystyle{icml2016}

\newpage
\begin{appendices}

\section{t-SNE Visualisation of Document Representations }
\label{app:a}
\begin{figure}[!h]
\centering
\subfloat[Neural Variational Document Model]{
  \centering
	\includegraphics[width=3in]{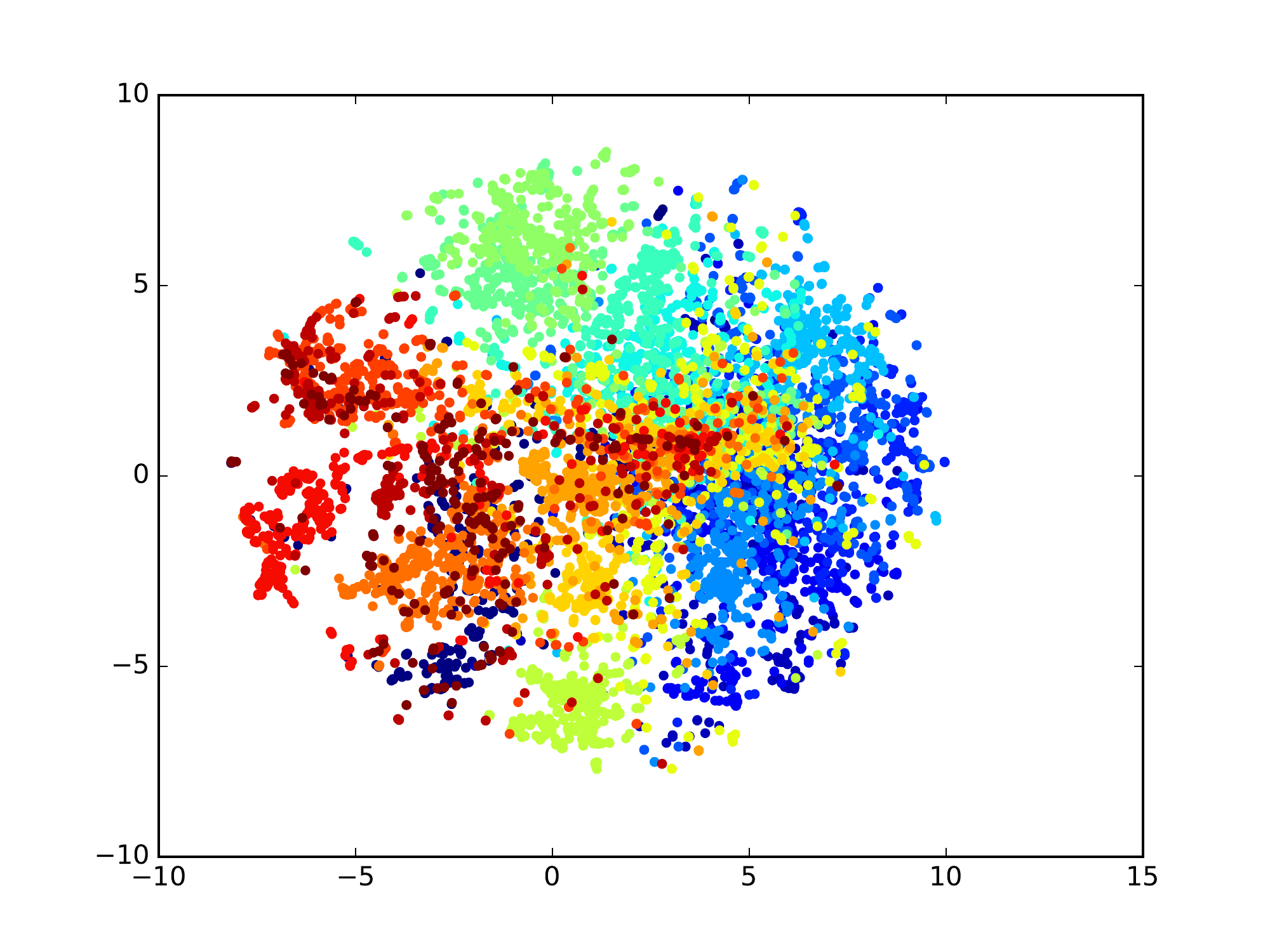}
}
\hfill
\subfloat[Semantic Word Vector]{
  \centering
	\includegraphics[width=3in]{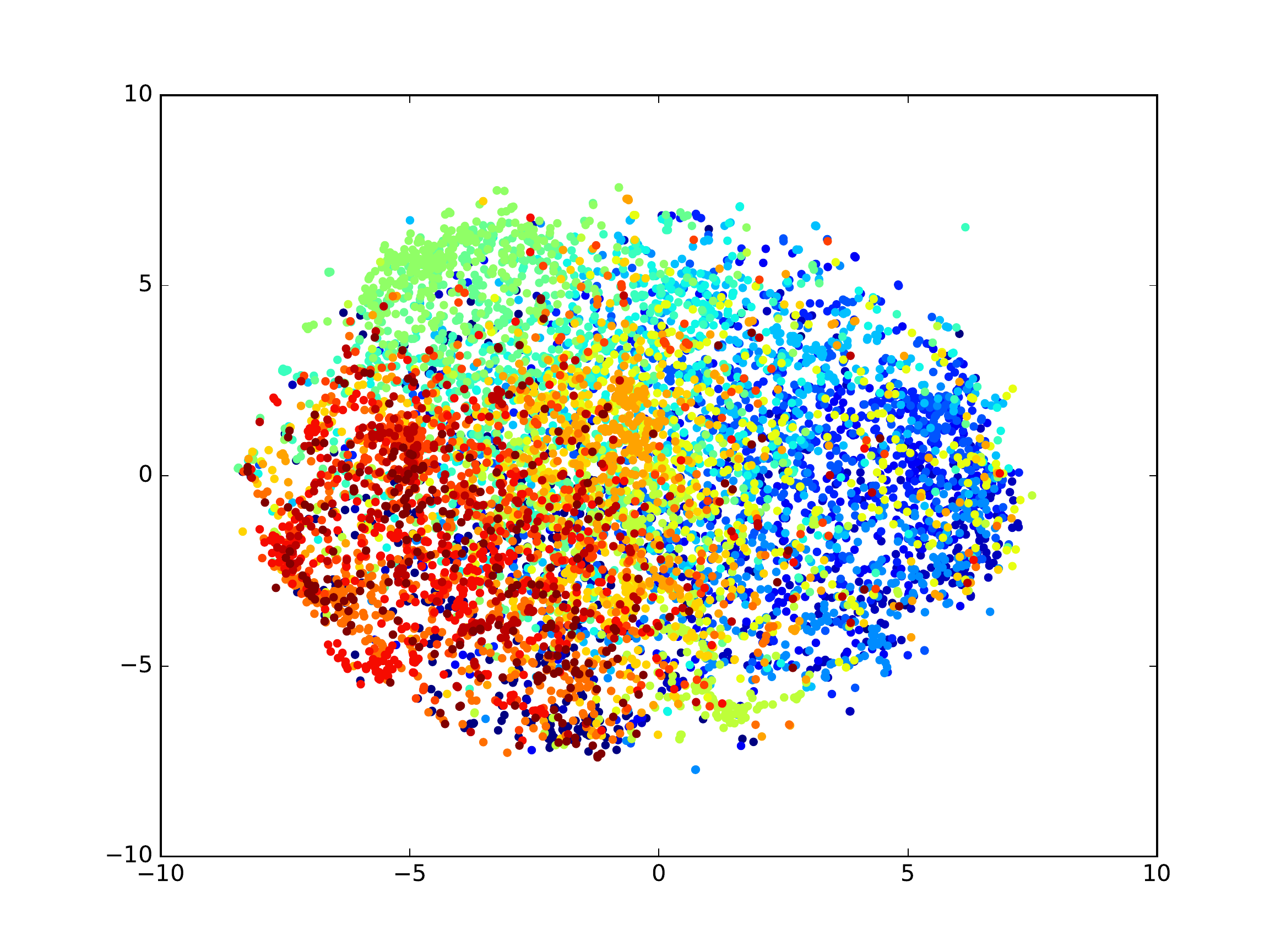}
}
\caption{t-SNE visualisation of the document representations achieved by (a) NVDM and (b) SWV \citep{maas2011learning} on the held-out test dataset of \textit{20NewsGroups}. The documents are collected from 20 different news groups, which correspond to the points with different colour in the figure. }
\end{figure}

\section{Details of the Deep Neural Network Structures }

\subsection{Neural Variational Document Model}
\label{app:nvdm}
(1) Inference Network $q_{\phi} ( \tmmathbf{h} | \tmmathbf{X} \nobracket)$:
\begin{eqnarray}
  & \tmmathbf{\lambda}_{} = \tmop{ReLU} ( \tmmathbf{W}_1
  \tmmathbf{X}+\tmmathbf{b}_1)  & \\
  & \tmmathbf{\pi} = \tmop{ReLU} ( \tmmathbf{W}_2 \tmmathbf{\lambda}_{}
  +\tmmathbf{b}_2) & \\
  & \tmmathbf{\mu} =\tmmathbf{W}_3 \tmmathbf{\pi} +\tmmathbf{b}_3  & \\
  & \log \tmmathbf{\sigma} =\tmmathbf{W}_4 \tmmathbf{\pi} +\tmmathbf{b}_4  &
  \\
  & \tmmathbf{h} \sim \mathcal{N} ( \tmmathbf{\mu} ( \tmmathbf{X}),
  \tmop{diag} ( \tmmathbf{\sigma}^2 ( \tmmathbf{X}))) &   
\end{eqnarray}

(2) Generative Model $p_{\theta} ( \tmmathbf{X} | \tmmathbf{h} \nobracket)$:
\begin{eqnarray}
  & \tmmathbf{\tmmathbf{e}}_i = \exp ( -\tmmathbf{h}^T
  \tmmathbf{R}\tmmathbf{x}_i +\tmmathbf{b}_{x_i})  & \\
  & p_{\theta} ( \tmmathbf{x}_i | \tmmathbf{h} \nobracket) =
  \frac{\tmmathbf{e}_i}{\sum^{| V |}_j \tmmathbf{e}_j} & \\
  & p_{\theta} ( \tmmathbf{X} | \tmmathbf{h} \nobracket) = \prod_i^N
  p_{\theta} ( \tmmathbf{x}_i | \tmmathbf{h} \nobracket) & 
\end{eqnarray}

(3) KL Divergence $D_{\tmop{KL}} [ q_{\phi} ( \tmmathbf{h} | \tmmathbf{X}
\nobracket) | | p ( \tmmathbf{h})]$:
\begin{eqnarray}
  & D_{\tmop{KL}} = - \frac{1}{2}  (K - \| \tmmathbf{\mu}_{} \|^2 - \| 
  \tmmathbf{\sigma} \|^2 + \log | \tmop{diag} ( \tmmathbf{\sigma}^2) |) & 
\end{eqnarray}

The variational lower bound to be optimised:
\begin{align}
  \nonumber \mathcal{L} = & \mathbbm{E}_{q_{\phi} ( \tmmathbf{h} | \tmmathbf{X}
  \nobracket)} \left[ \sum\nolimits_{i = 1}^N \log p_{\theta} ( \tmmathbf{x}_i |
  \tmmathbf{h} \nobracket) \right] \\
  &- D_{\tmop{KL}} [ q_{\phi} ( \tmmathbf{h} |
  \tmmathbf{X} \nobracket) | | p ( \tmmathbf{h})]\\
  \nonumber \thickapprox & \sum\nolimits_{l = 1}^L \sum\nolimits_{i = 1}^N \log p_{\theta} (
  \tmmathbf{x}_i | \tmmathbf{h} \nobracket^{( l)})  \\
  & + \frac{1}{2}  (K - \|
  \tmmathbf{\mu}_{} \|^2 - \|  \tmmathbf{\sigma} \|^2 + \log | \tmop{diag} (
  \tmmathbf{\sigma}^2) |)
\end{align}

\subsection{Neural Answer Selection Model}
\label{app:nasm}
(1) Inference Network $q_{\phi} ( \tmmathbf{h} | \tmmathbf{q}, \tmmathbf{a},
\tmmathbf{y} \nobracket)$:
\begin{eqnarray}
  & \tmmathbf{s}_q ( | \tmmathbf{q} |) = f_q^{\textsc{lstm}} ( \tmmathbf{q}) & 
   \\
  & \tmmathbf{s}_a ( | \tmmathbf{a} |) = f_a^{\textsc{lstm}} ( \tmmathbf{a}) & 
   \\
  & \tmmathbf{s}_y =\tmmathbf{W}_5 \tmmathbf{y}+\tmmathbf{b}_5 &   \\
  & \tmmathbf{\gamma}=\tmmathbf{s}_q ( | \tmmathbf{q} |) | | \tmmathbf{s}_a (
  | \tmmathbf{a} |) | | \tmmathbf{s}_y &   \\
  & \tmmathbf{\lambda}_{\phi} = \tanh ( \tmmathbf{W}_6
  \tmmathbf{\gamma}+\tmmathbf{b}_6) &   \\
  & \tmmathbf{\pi}_{\phi} = \tanh ( \tmmathbf{W}_7
  \tmmathbf{\tmmathbf{\lambda}_{\phi}} +\tmmathbf{b}_7) &   \\
  & \tmmathbf{\mu}_{\phi} =\tmmathbf{W}_8 \tmmathbf{\tmmathbf{\pi}_{\phi}}
  +\tmmathbf{b}_8  &   \\
  & \log \tmmathbf{\sigma}_{\phi} =\tmmathbf{W}_9
  \tmmathbf{\tmmathbf{\pi}_{\phi}} +\tmmathbf{b}_9  &   \\
  & \tmmathbf{h} \sim \mathcal{N} ( \tmmathbf{\mu}_{\phi} ( \tmmathbf{q},
  \tmmathbf{a}, \tmmathbf{y}), \tmop{diag} ( \tmmathbf{\sigma}_{\phi}^2 (
  \tmmathbf{q}, \tmmathbf{a}, \tmmathbf{y}))) &   
\end{eqnarray}

(2) Generative Model

$p_{\theta} ( \tmmathbf{h} | \tmmathbf{q} \nobracket)$:
\begin{eqnarray}
  & \tmmathbf{\lambda}_{\theta} = \tanh ( \tmmathbf{W}_1 \tmmathbf{s}_q ( |
  \tmmathbf{q} |) +\tmmathbf{b}_1)  &   \\
  & \tmmathbf{\pi}_{\theta} = \tanh ( \tmmathbf{W}_2
  \tmmathbf{\lambda}_{\theta} +\tmmathbf{b}_2) &   \\
  & \tmmathbf{\mu}_{\theta} =\tmmathbf{W}_3 \tmmathbf{\pi}_{\theta}
  +\tmmathbf{b}_3  &   \\
  & \log \tmmathbf{\sigma}_{\theta} =\tmmathbf{W}_4
  \tmmathbf{\pi_{}}_{\theta} +\tmmathbf{b}_4  &   
\end{eqnarray}
$p_{\theta} ( \tmmathbf{y} | \tmmathbf{q}, \tmmathbf{a}, \tmmathbf{h}
\nobracket)$:
\begin{eqnarray}
  & \tmmathbf{e} ( i) =\tmmathbf{W}_{\alpha}^T \tanh ( \tmmathbf{W}_h
  \tmmathbf{h}+\tmmathbf{W}_s \tmmathbf{s}_a ( i)) &   \\
  & \alpha ( i) = \frac{\tmmathbf{e} ( i)}{\sum_j \tmmathbf{e} ( j)} & 
   \\
  & \tmmathbf{c} ( \tmmathbf{a}, \tmmathbf{h}) = \sum_i \tmmathbf{s}_a ( i)
  \alpha ( i) &   \\
  & \tmmathbf{z}_a ( \tmmathbf{a}, \tmmathbf{h}) = \tanh ( \tmmathbf{W}_a
  \tmmathbf{c} ( \tmmathbf{a}, \tmmathbf{h}) +\tmmathbf{W}_n \tmmathbf{s}_a (
  | \tmmathbf{a} |)) &   \\
  & \tmmathbf{z}_q ( \tmmathbf{q}) =\tmmathbf{s}_q ( | \tmmathbf{q} |) & 
   \\
  & p_{\theta} ( \tmmathbf{y}= 1 | \tmmathbf{q}, \tmmathbf{a}, \tmmathbf{h}
  \nobracket) = \sigma ( \tmmathbf{z}_q^T \tmmathbf{M}\tmmathbf{z}_a + b) & 
\end{eqnarray}

(3) KL Divergence $D_{\tmop{KL}} [ q_{\phi} ( \tmmathbf{h} | \tmmathbf{q},
\tmmathbf{a}, \tmmathbf{y} \nobracket) | | p_{\theta} ( \tmmathbf{h} |
\tmmathbf{q} \nobracket)]$:
\begin{align}
  \nonumber D_{\tmop{KL}}  = & - \frac{1}{2}  (K + \log | \tmop{diag} (
  \tmmathbf{\sigma}_{\phi}^2) | - \log | \tmop{diag} (
  \tmmathbf{\sigma}_{\theta}^2) | \\
  \nonumber & -  \tmop{Tr} ( \tmop{diag} (
  \tmmathbf{\sigma}^2_{\phi}) \tmop{diag}^{-1} (
  \tmmathbf{\sigma}^2_{\theta}))  \\ 
  & - ( \tmmathbf{\mu}_{\phi} -
  \tmmathbf{\mu}_{\theta})^T \tmop{diag}^{-1} (
  \tmmathbf{\sigma}^2_{\theta}) ( \tmmathbf{\mu}_{\phi} -
  \tmmathbf{\mu}_{\theta}))
\end{align}

The variational lower bound to be optimised:
\begin{align}
  \nonumber \mathcal{L}  = & \  \mathbbm{E}_{q_{\phi} ( \tmmathbf{h} | \tmmathbf{q},
  \tmmathbf{a}, \tmmathbf{y} \nobracket)} [ \log p_{\theta} ( \tmmathbf{y} |
  \tmmathbf{q}, \tmmathbf{a}, \tmmathbf{h} \nobracket)] \\
  & - D_{\tmop{KL}} [q_{\phi} ( \tmmathbf{h} | \tmmathbf{q}, \tmmathbf{a}, \tmmathbf{y}
  \nobracket) | | p_{\theta} ( \tmmathbf{h} | \tmmathbf{q} )]\\
  \nonumber \thickapprox & \sum_{l = 1}^L [ \tmmathbf{y} \log \sigma ( \tmmathbf{z}_q^T
  \tmmathbf{M}\tmmathbf{z}_a^{(l)} + b) \\
  \nonumber & + ( 1 -\tmmathbf{y}) \log ( 1 - \sigma (
  \tmmathbf{z}_q^T \tmmathbf{M}\tmmathbf{z}_a^{(l)} + b)) ] \\
  \nonumber &   + \frac{1}{2}  (K + \log | \tmop{diag} ( \tmmathbf{\sigma}_{\phi}^2) | -
  \log | \tmop{diag} ( \tmmathbf{\sigma}_{\theta}^2) | \\
  & - \tmop{Tr} (
  \tmop{diag}^{} ( \tmmathbf{\sigma^{}}^2_{\phi}) \tmop{diag}^{- 1} (
  \tmmathbf{\sigma^{}}^2_{\theta}))  \nonumber \\
  & - ( \tmmathbf{\mu}_{\phi} -
  \tmmathbf{\mu}_{\theta})^T \tmop{diag}^{- 1} (
  \tmmathbf{\sigma^{}}^2_{\theta}) ( \tmmathbf{\mu}_{\phi} -
  \tmmathbf{\mu}_{\theta}))
\end{align}

\newpage

\section{Computational Complexity}
\label{app:c}
The computational complexity of NVDM for a training document is $C_\phi+C_\theta=O(LK^2+KSV)$. Here, $C_\phi=O(LK^2)$ represents the cost for the inference network to generate a sample, where $L$ is the number of the layers in the inference network and $K$ is the average dimension of these layers. Besides, $C_\theta=O(KSV)$ is the cost of reconstructing the document from a sample, where $S$ is the average length of the documents and $V$ represents the volume of words applied in this document model, which is conventionally much lager than $K$.

The computational complexity of NASM for a training question-answer pair is $C_\phi+C_\theta=O((L+S)K^2+SW)$. 
The inference network needs $C_\phi=2SW+2K+LK^2=O(LK^2+SW)$.
It takes $2SW+2K$ to produce the joint representation for a question-answer pair and its label, where $W$ is the total  number of parameters of an LSTM and $S$ is the average length of the sentences. Based on the joint representation, an MLP spends $LK^2$ to generate a sample, where $L$ is the number of layers and $K$ represents the average dimension. 
The generative model requires  $C_\theta=2SW+LK^2+SK^2+5K^2+2K^2=O((L+S)K^2+SW)$. 
Similarly, it costs $2SW+LK^2$ to construct the generative latent distribution , where $2SW$ can be saved if the LSTMs are shared by the inference network and the generative model. Besides, the attention model takes $SK^2+5K^2$ and the relatedness prediction takes the last $2K^2$. 

Since the computations of NVDM and NASM can be parallelised in GPU and only one sample is required during training process, it is very efficient to carry out the neural variational inference. As NVDM is an instantiation of variational auto-encoder, its computational complexity is the same as the deterministic auto-encoder. In addition, the computational complexity of LSTM+Att, the deterministic counterpart of NASM, is also $O((L+S)K^2+SW)$. There is only $O(LK^2)$ time increase by introducing an inference network for NASM when compared to LSTM+Att.

\end{appendices}

\end{document}